\documentclass{article}

\usepackage{arxiv}

\usepackage[utf8]{inputenc} 
\usepackage[T1]{fontenc}    
\usepackage{hyperref}       
\usepackage{url}            
\usepackage{booktabs}       
\usepackage{amsfonts}       
\usepackage{nicefrac}       
\usepackage{microtype}      
\usepackage{lipsum}		
\usepackage{amsthm}
\usepackage{graphicx}
\usepackage{doi}
\usepackage{physics}
\usepackage{amssymb}
\usepackage{pifont}

\newcommand{\pd}[2]{\frac{\partial #1}{\partial #2}}
\newcommand{\pdd}[2]{\frac{\partial^2 #1}{\partial #2^2}}
\newcommand{\bs}{\boldsymbol}
\newcommand{\mbf}{\mathbf}

\usepackage{setspace}
\title{Visualizing the loss landscapes of physics-informed neural networks}

\date{} 					

\author{
    Conor Rowan \\
	Aerospace Engineering\\
	University of Colorado Boulder\\
    3775 Discovery Drive \\
	Boulder, CO 80309 \\
	\texttt{conor.rowan@colorado.edu} \\
    \And  
    Finn Murphy-Blanchard \\
	Applied Mathematics\\
	University of Colorado Boulder\\
    1111 Engineering Drive\\
	Boulder, CO 80310 \\
	\texttt{finn.murphy-blanchard@colorado.edu} \\
}

\renewcommand{\headeright}{}
\renewcommand{\undertitle}{}
\renewcommand{\shorttitle}{}

\begin{document}
\maketitle

\begin{abstract}
    Training a neural network requires navigating a high-dimensional, non-convex loss surface to find parameters that minimize this loss. In many ways, it is surprising that optimizers such as stochastic gradient descent and ADAM can reliably locate minima which perform well on both the training and test data. To understand the success of training, a ``loss landscape'' community has emerged to study the geometry of the loss function and the dynamics of optimization, often using visualization techniques. However, these loss landscape studies have mostly been limited to machine learning for image classification. In the newer field of physics-informed machine learning, little work has been conducted to visualize the landscapes of losses defined not by regression to large data sets, but by differential operators acting on state fields discretized by neural networks. In this work, we provide a comprehensive review of the loss landscape literature, as well as a discussion of the few existing physics-informed works which investigate the loss landscape. We then use a number of the techniques we survey to empirically investigate the landscapes defined by the Deep Ritz and squared residual forms of the physics loss function. We find that the loss landscapes of physics-informed neural networks have many of the same properties as the data-driven classification problems studied in the literature. Unexpectedly, we find that the two formulations of the physics loss often give rise to similar landscapes, which appear smooth, well-conditioned, and convex in the vicinity of the solution. The purpose of this work is to introduce the loss landscape perspective to the scientific machine learning community, compare the Deep Ritz and the strong form losses, and to challenge prevailing intuitions about the complexity of the loss landscapes of physics-informed networks.
\end{abstract}



\tableofcontents

\section{Introduction}
\label{intro}

\paragraph{} In machine learning, the loss function defines a high-dimensional surface on the space of neural network parameters. Training the network requires finding a minimum of this non-convex surface through iterative gradient-based optimization strategies. The proven ability of optimizers such as stochastic gradient descent (SGD) and ADAM to successfully navigate these complex surfaces is surprising and counterintuitive. Geoffrey Hinton, the Nobel Laureate and so-called ``Godfather of AI,'' said in the 1980s that neural networks were ``not worth investigating'' since one is ``bound to get stuck in local minima.'' To this day, it remains mysterious that, in general, local minima do not seriously hamper network training and that trained networks generalize to unseen data. Despite the incredible successes of machine learning (ML) in the last decade, each perplexing success is accompanied by an equally-perplexing failure of the training process. For example, the spectral bias of neural networks makes fitting high-frequency behavior challenging \cite{rahaman_spectral_2019}, which can cause even one-dimensional regression problems to fail. How can neural networks learn to classify images, or generate novel artistic works, yet struggle to represent an oscillatory univariate function? Much like engineers of the 18th and 19th centuries who built engines without a robust theory of thermodynamics, we have harnessed the power of neural networks without understanding precisely how they learn.

\paragraph{} One avenue of insight into neural network training is the loss landscape. Studying mostly data-driven image classification problems, a coherent ``loss landscape'' research community has formed, which uses a variety of mathematical and computational techniques to better understand the training process. Comprising empiricists with backgrounds in computer science and engineering, mathematicians, and statistical physicists, this community has helped clarify why training is often successful, and has uncovered a number of intriguing features of neural network loss landscapes. In our review, we divide loss landscape research into three broad subcategories: mathematics, statistical physics, and empirics. Each of these subcategories is reviewed below. We then discuss physics-informed machine learning, whose loss landscapes will be the focus of this work.

\subsection{Mathematics} 

\paragraph{} Generally speaking, mathematical approaches to studying the loss landscape prove general properties of the training process and of critical points. For example, the exact training dynamics for deep linear networks were ascertained in \cite{saxe_exact_2014}. The neural tangent kernel (NTK) proved that the training of even deep nonlinear networks can be treated as a dynamical system with exact solutions, so long as the network is sufficiently wide \cite{jacot_neural_2020}. In \cite{fukumizu_local_2000}, it was demonstrated that networks have a hierarchical structure, whereby a minium of a subnetwork is guaranteed to be a critical point of a larger network. \cite{simsek_geometry_2021} identified and characterized ``permutation invariances'' in multilayer perceptron networks, which guarantee that the loss landscape has many local minima of exactly equal loss value. These critical points have been shown to be connected by paths over which the training loss does not increase \cite{brea_weight-space_2019}. This reflects the findings of \cite{cooper_loss_2018}, which proves that there are connected manifolds of parameters which minimize the loss. In \cite{du_gradient_2019}, it was proven that gradient descent finds zero loss solutions when the network is overparameterized. Finally, a number of works have demonstrated the surprising property that there are no ``bad minima'' when the network is overparameterized, meaning that, under certain assumptions, all local minima are also global minima \cite{kawaguchi_deep_2016, laurent_deep_2018, nguyen_loss_2017, nguyen_loss_2018, sun_global_2020}. This result casts light on why non-convexity of the loss landscape need not adversely affect training. However, \cite{swirszcz_local_2017} demonstrates that some problems of practical interest do have poor-performing local minima.

\subsection{Statistical physics}

\paragraph{} At first glance, neural network training seems far removed from the stochastic microscopic systems typically studied in statistical physics. However, \cite{mandt_stochastic_2018} showed that SGD, a common strategy for training data-driven classifiers, could be interpreted as solving a stochastic differential equation with additive noise. In this case, the density of the parameters follows a Fokker-Planck equation whose steady state is a Gibbs distribution. Here, the role of energy is played by the loss function and the temperature is a measure of the SGD noise. Thus, with the neural network parameters as the ``microstate'' and the loss value as the ``macrostate,'' a strong connection from neural network training to statistical physics was established. For example, noting that ``flat'' minima of the loss often tend to generalize better to unseen data, some authors have used the idea of statistical entropy to design optimizers which strategically target these flatter minima \cite{chaudhari_entropy-sgd_2017, pittorino_entropic_2021}. The concept of a phase transition has also been fruitful in the study of neural networks, as it has been shown that the network experiences rapid changes in behavior as hyperparameters such as the learning rate and the batch size are varied \cite{yang_taxonomizing_2021, geiger_jamming_2019, geiger_landscape_2021, nakhodnov_loss_2022}. Statistical physicists have also used their experience with the non-convex loss landscapes defined by spin glass systems such as the Ising model to explore neural network loss landscapes \cite{li_spin_2025, sagun_explorations_2015, liao_exploring_2024, barra_how_2012, choromanska_loss_2015}. Finally, statistical learning theory, which borrows tools from statistical physics, can be used to understand the implicit biases of different optimizers, which helps explain the properties of solutions obtained in practice \cite{gerbelot_applying_2024}.

\subsection{Empirics}

\paragraph{} The final pillar of loss landscape research is empirical studies. These works do not claim to uncover universal properties of the loss landscape, but rather they make observations about the landscapes encountered in particular problems. Perhaps the most obvious empirical approach is to directly visualize the loss landscape. One strategy from topological data analysis is to use ``merge trees'' to visualize the connectivity of minima \cite{li_sketching_2021}, but this approach has not been popular for machine learning. Because of the large dimensionality of these problems, directly visualizing the loss surface requires taking cross-sections in user-specified lines or planes. The engineer and entrepreneur Javier Ideami has made these visualizations an artistic venture, showcasing the precipitous landscapes of modern machine learning models in impressive detail \cite{ideami_loss_2026}. \cite{im_empirical_2017} chose to visualize the loss in the plane spanned by randomly initialized parameters and the solutions obtained from two different optimizers. \cite{li_visualizing_2018} used random directions in parameter space to construct the cross-sectional contour plots. Another popular approach has been to use eigenvectors of the Hessian of the loss to define cross-sections, an approach which is pursued in \cite{bottcher_visualizing_2024}. In \cite{xu_visualizing_2025}, the loss was plotted along a line in parameter space as opposed to a plane, and different shapes of the loss profile were taxonomized.

\paragraph{} Apart from building contour plots, the Hessian matrix of the loss has proven to be a useful tool for understanding the loss landscape. For machine learning problems, the majority of the Hessian eigenvalues concentrate around zero, with a small set indicating large, positive curvature \cite{sagun_eigenvalues_2017, sagun_empirical_2018, ghorbani_investigation_2019}. In fact, the Hessian frequently has negative eigenvalues \cite{alain_negative_2019}, which corroborates the claim that saddle points outnumber minima in high dimensions \cite{dauphin_identifying_2014}. Visualizing the evolution of Hessian eigenvalues over the course of training can provide insight into the geometry of the basins the optimizer finds \cite{yao_pyhessian_2020}. Finally, \cite{fort_goldilocks_2018} found more convex regions of the loss at certain radii in parameter space, which they called the ``Goldilocks zone.''

\paragraph{} Another approach to investigate the loss landscape is dimensionality reduction. For example, it has been shown that networks can train effectively in low-dimensional random affine subspaces \cite{li_measuring_2018}. The authors used the dimension of this subspace to determine the ``intrinsic dimensionality'' of the network. In a similar spirit, the ``lottery ticket hypothesis'' suggests that specific subnetworks of larger networks can obtain similar train and test performance, which again hints at an intrinsic low-dimensional network structure \cite{frankle_lottery_2019}. \cite{horoi_exploring_2022} uses nonlinear dimensionality reduction techniques on data generated from gradient descent to visualize the difference between networks that generalize well and those that don't. In many cases, the dynamics of the network parameters are confined to a low-dimensional subspace, which aids in visualizing the training process \cite{gallagher_visualization_2003}. Other authors found that this subspace was spanned by the eigenvectors of the Hessian corresponding to a handful of the largest eigenvalues \cite{gur-ari_gradient_2018}. An additional dimensionality reduction approach is to visualize the loss along a path connecting the initial and final parameters. It has been demonstrated empirically that the loss often decreases monotonically along this path, a phenomenon that has been named ``monotonic linear interpolation'' (MLI) \cite{goodfellow_qualitatively_2015, wang_plateau_2023}. While MLI appears frequently, \cite{lucas_analyzing_2021} showed that it can be systematically violated by adjusting the optimizer and learning rate.

\paragraph{} ``Mode connectivity'' occurs when two different solutions are connected by a path in parameter space over which the loss does not increase. This has been another robust empirical finding in the loss landscape literature, suggesting that solutions are not isolated minima, rather they are basins of equivalent loss value \cite{draxler_essentially_2019, fort_large_2019, garipov_loss_2018}. This is related to the observations of \cite{lipton_stuck_2016}, which suggest that solutions obtained from training are not strict local minima; rather they are vast flat regions. Mode connectivity has been shown to be robust to different training schemes and initializations \cite{gotmare_using_2018}, and has also been explored in light of the permutation invariance of networks \cite{entezari_role_2022}. It has also been studied in the setting of convex optimization by considering the dual of the original training problem \cite{kim_exploring_2025}.

\paragraph{} Finally, we note that \cite{zhang_understanding_2017} illustrated that neural networks can be trained to zero loss on classification problems when the class labels are completely random, even with regularization on the norm of the parameters. This shows that the network is expressive enough to fit data with no meaningful structure, and thus, that the model architecture alone does not explain inductive biases responsible for generalization. Yet, \cite{huang_understanding_2020} showed that a deep network classifier significantly outperformed a linear model on the test data, even when both obtained zero error in training. The authors argued that despite the presence of solutions with poor generalization properties in the neural network loss landscape, the optimizer favored the ``good'' solution because it has higher volume in the loss landscape. The larger volume of the generalizable solution is a consequence of the smaller curvature of the loss about this point, which suggests that the performance of the classifier is locally insensitive to changes in the network parameters. The authors argue that this insensitivity is a consequence of decision boundaries which are maximally distant from the data, and, in some sense, more parsimonious. We speculate that this notion of high-volume solutions can be connected to the spectral bias of neural networks \cite{rahaman_spectral_2019}, sometimes referred to as the ``F-principle'' \cite{xu_frequency_2020}, though this connection has yet to be made in the literature.

\subsection{Physics-informed machine learning}

\paragraph{} As the above review shows, loss landscapes have been thoroughly investigated in the literature on machine learning for data-driven classification problems. The vast majority of the empirical works we have discussed train networks for image classification based on the MNIST and/or CIFAR-10 data sets. However, parallel to classification problems in ML, physics-informed machine learning has established itself as a successful and independent area of research, with wide applicability to computational science and engineering. Unlike mainstream machine learning, physics-informed training is often supervised exclusively by the governing ordinary or partial differential equations, and thus need not rely on data. While the loss landscapes of physics-informed problems are still defined over the space of neural network parameters, the loss function itself differs drastically from standard mean-squared error or cross-entropy losses used in classification problems. In the original physics-informed neural networks (PINNs) framework, the loss function consists of the squared residual of the governing partial differential equation (PDE) with penalties to enforce boundary conditions \cite{raissi_physics-informed_2019, sirignano_dgm_2018}. While the norm of the weak form system represents another choice of loss function \cite{khodayi-mehr_varnet_2019}, we view the variational energy, if it exists, as a more natural and elegant alternative \cite{e_deep_2017}. With the solution field discretized by a neural network, minimizing the variational energy is called the ``Deep Ritz method'' (DRM), and is beneficial because it lowers the order of differentiation and incorporates Neumann boundary conditions without penalties. Further modifications to the loss function can be made by adjusting the method of boundary condition enforcement, such as by hard-coding the Dirichlet boundaries \cite{sukumar_exact_2022} or using the Augmented Lagrangian method \cite{son_enhanced_2023}.

\paragraph{} Very few investigations of the physics-informed loss landscape have been conducted. \cite{gopakumar_loss_2023} argued that augmenting the physics-informed objective with measurement data leads to sharper minima using surface plots of cross-sections of the loss function. A roughness index of physics-informed objectives was introduced in \cite{wu_roughness_2023}, which provides a measure of the complexity of the loss landscape. The authors showed contour plots of cross-sections of the loss landscapes from both PINNs and DRM objectives. A number of works have used the Hessian to show that physics-informed loss landscapes are frequently ill-conditioned, indicating that preconditioning from second-order optimizers is beneficial \cite{cao_analysis_2025, rathore_challenges_2024, krishnapriyan_characterizing_2021}. The neural tangent kernel has also been explored in physics-informed settings, leading to modifications of the network architecture to mitigate the spectral bias \cite{wang_eigenvector_2021}. The NTK has also been used to diagnose training failures in networks by showing that different terms in the loss function have different convergence rates \cite{wang_when_2022}.

\subsection{This work}

\paragraph{} While many of these works touch on aspects of loss landscape visualization, no systematic loss landscape studies have been conducted in the physics-informed machine learning literature. In other words, there are myriad tools that have been developed in the broader loss landscape community that have yet to be explored in the physics-informed context. Our intention is to introduce many of these existing techniques for empirical analysis of the loss landscape into the PINNs community. This means that we will not pursue the mathematical or statistical mechanics approaches any further. Using two example problems, we study whether physics-informed neural networks have the properties of 1) monotonic linear interpolation, 2) mode connectivity, 3) Hessian singularity, 4) a preferred parameter radius, i.e., a ``Goldilocks zone,'' 5) low intrinsic dimensionality, 6) low-dimensional optimization dynamics, and 7) no bad local minima. Along the way, we introduce the novel concepts of a ``Hessian walk,'' used to explore isocontours of the loss surface, and the ``acceleration'' of the parameters, which we take as a measure of curving/bending in the loss landscape. We also decompose the network into coefficients and basis functions, which allows us to interpret one source of Hessian singularity. By constructing neural network discretizations that automatically satisfy the boundary conditions, we intentionally avoid the confounding effect of boundary condition enforcement in the loss landscape. The purpose of this study is to visualize loss landscapes arising purely from the governing equation, and to compare the landscapes obtained from two distinct notions of a solution to the governing equation.


\section{One-dimensional elliptic problem}
\label{1d}

\paragraph{} The first problem whose loss landscape we study is a one-dimensional elliptic boundary value problem with homogeneous Dirichlet boundary conditions. The governing equation is 

\begin{equation}\label{1d_govern}
    \pdd{u}{x}+f(x) =0, \quad u(0)=u(1)=0,
\end{equation}

\noindent where $x\in[0,1]$ is the spatial coordinate, $u(x)$ is a state field such as temperature or displacement, and $f(x)$ is the source term. The problem will be studied from the perspective of the standard PINNs approach relying on the strong form loss \cite{raissi_physics-informed_2019} and that of the Deep Ritz method using the variational energy \cite{e_deep_2017}. In both cases, the state field is discretized as

\begin{equation*}
    u(x;\bs \theta) = \sin(\pi x) \mathcal N(x; \bs \theta),
\end{equation*}

\noindent where $\mathcal N: \mathbb R \rightarrow \mathbb R$ is a multilayer perceptron neural network with parameters $\bs \theta$ and $\sin(\pi x)$ is a distance-type function enforcing the homogeneous Dirichlet boundary conditions \cite{sukumar_exact_2022}\footnote{The subsequent subsection on the Hessian eigenvalue evolution shows that we actually enforce the boundary conditions on each of the neurons in the final layer of the network, which we think of as the basis functions. The current presentation is more standard and avoids presenting details which are unnecessary at this point.}. We denote the number of trainable network parameters with $|\bs \theta|$. A solution to Eq. \eqref{1d_govern} with the Deep Ritz method is given by 

\begin{equation}\label{drm_1d}
    \Pi(\bs \theta) = \int_0^1 \frac{1}{2}\qty( \pd{u(x;\bs\theta)}{x})^2 - f(x) u(x; \bs \theta) dx, \quad \bs \theta_f^{\text{DRM}} = \underset{\bs \theta}{\text{argmin }} \Pi(\bs \theta).
\end{equation}

The randomly initialized parameters used as the starting point of the optimization problem in Eq. \eqref{drm_1d} are denoted $\bs \theta^{\text{DRM}}_i$. Going forward, we call $\Pi: \mathbb R^{| \bs\theta|} \rightarrow \mathbb R$ the ``energy'' or the DRM objective. We remark that we use the terms objective and loss interchangeably. A solution to Eq. \eqref{1d_govern} with a PINN (squared residual) approach is given by

\begin{equation}\label{sf_1d}
    \mathcal L(\bs \theta) = \frac{1}{2}\int_0^1 \qty( \pdd{u(x;\bs \theta)}{x} + f(x) )^2 dx, \quad \bs \theta^{\text{PINN}}_f = \underset{\bs \theta}{\text{argmin }} \mathcal L(\bs \theta).
\end{equation}

The randomly initialized parameters used as the starting point of the optimization problem in Eq. \eqref{sf_1d} are denoted $\bs \theta^{\text{PINN}}_i$. Going forward, we call $\mathcal L: \mathbb R^{|\theta|} \rightarrow \mathbb R$ the ``strong form'' or PINN objective. The energy and strong form loss objectives both define landscapes whose minima correspond to solutions of the given boundary value problem. In all following numerical examples, we assume a manufactured solution of $u(x)=-2x\sin(2\pi x)$ and compute the source term as $f(x) = -\partial^2 u / \partial x^2$. The integration of the DRM and PINN objectives is accomplished with $100$ uniformly spaced integration points on a fixed grid. The details of the optimization and network architecture will be specified on a case-by-case basis. However, we choose hyperbolic tangent activation functions for all networks. In the following subsections, we study these loss landscapes with techniques from the literature as well as novel techniques which are tailored to the physics-informed setting.

\subsection{Monotonic linear interpolation}

\paragraph{} The current literature on visualizing loss landscapes for machine learning problems is devoted almost entirely to data-driven image classification problems. A number of authors have noted that in these problems, the objective function monotonically decreases along a straight-line path connecting the initial and final parameters \cite{goodfellow_qualitatively_2015, wang_plateau_2023}. This property is known as ``monotonic linear interpolation'' (MLI). We note that, while MLI has been demonstrated empirically a number of times, networks can be designed to systematically violate it \cite{lucas_analyzing_2021}. Here, we test whether the MLI property holds for the Deep Ritz and PINN objective functions. The state field is discretized with a two hidden-layer network of width $20$ and the optimization problems in Eqs. \eqref{drm_1d} and \eqref{sf_1d} are solved using ADAM optimization for $7500$ epochs with a learning rate of $1 \times 10^{-3}$. We confirm that both methods converge to the true solution with the given discretization and optimization settings. We are interested in whether the following two curves decrease monotonically with the interpolation parameter $t\in[0,1]$:

\begin{equation*}
    \Pi( \bs \theta_i^{\text{DRM}} + t( \bs\theta_f^{\text{DRM}} - \bs \theta^{\text{DRM}}_i) ), \quad \mathcal L( \bs \theta_i^{\text{PINN}} + t( \bs\theta_f^{\text{PINN}} - \bs \theta^{\text{PINN}}_i) ).
\end{equation*}

\begin{figure}[hbt!]
\centering
\includegraphics[width=0.99\textwidth]{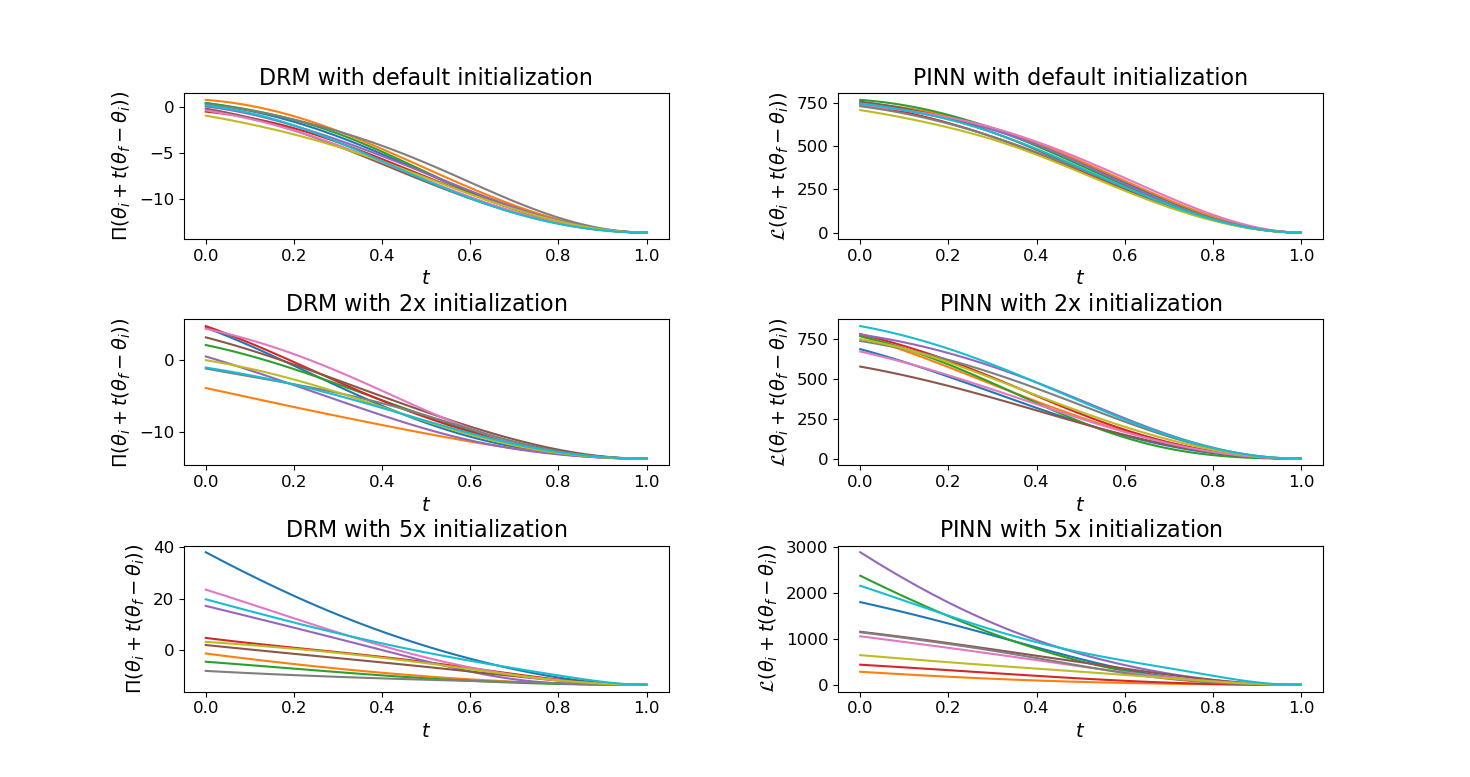}
\caption{Exploring the MLI property of the Deep Ritz and PINNs loss landscapes. Out of the $60$ trials, every one shows a monotonic decrease of the loss function along the straight-line path connecting the initial and final parameters.}
\label{MLI}
\end{figure}

To investigate both the existence and the robustness of the MLI property, we initialize the parameters for both the energy and strong form loss with three different initialization strategies. For the first, we use the default PyTorch ``Kaiming'' initialization for linear layers \cite{he_delving_2015}. The second approach is to take the default initialization and scale it up by a factor of $2$, whereas the third strategy does the same but with a factor of $5$. These experiments are meant to explore whether the MLI property is particular to certain radii in parameter space. Each initialization is repeated $10$ times. See Figure \ref{MLI} for the results. In all cases, the linear interpolation between the initial and final parameters shows a monotonic decrease in the loss.

\subsection{Exploring the solution manifold}

\paragraph{} Loss landscapes defined by neural networks are guaranteed to have many local minima with the same loss value given their so-called ``permutation invariance,'' which arises from the fact that parameters can be re-ordered without affecting the mathematical structure of the network \cite{simsek_geometry_2021}. Thus, a network trained from a random initialization need not obtain the same solution each time, with permutation invariance being but one source of this non-uniqueness. A surprising result from the literature is that different solutions which obtain equivalent loss values are connected by paths over which the loss does not increase. This is the phenomenon of mode connectivity, which points to the fact that the loss surface contains connected basins of minima \cite{draxler_essentially_2019, fort_large_2019, garipov_loss_2018, gotmare_using_2018}. In this example, we introduce a technique to explore these basins. To this end, we use the same network and optimization process to obtain solutions to the DRM and PINN problems, given by $\bs \theta^{\text{DRM}}_f$ and $\bs \theta^{\text{PINN}}_f$ respectively. For each set of parameters, we then perform a ``Hessian walk'' around the solution as follows:

\begin{equation}\label{hessian_walk}
    \bs \theta_t = \bs \theta_{t-1} + \eta \mbf v_{\text{min}}, \quad t=1,2,\dots,
\end{equation}

\noindent where $\mbf v_{\text{min}}$ is the eigenvector corresponding to the minimum absolute eigenvalue of the Hessian matrix and $\eta$ is a user-specified step size. The initial parameter of the Hessian walk is the converged solution from the physics-informed training, i.e., $\bs \theta_0=\bs \theta_f$. For the two formulations of the physics loss, these Hessian matrices are given by 

\begin{equation*}
    \mbf J^{\text{DRM}} = \frac{\partial \Pi}{\partial \bs \theta \partial \bs \theta}(\bs \theta^{\text{DRM}}_f), \quad \mbf J^{\text{PINN}} = \frac{\partial \mathcal L}{\partial \bs \theta \partial \bs \theta}(\bs \theta^{\text{PINN}}_f).
\end{equation*}

\begin{figure}[hbt!]
\centering
\includegraphics[width=0.99\textwidth]{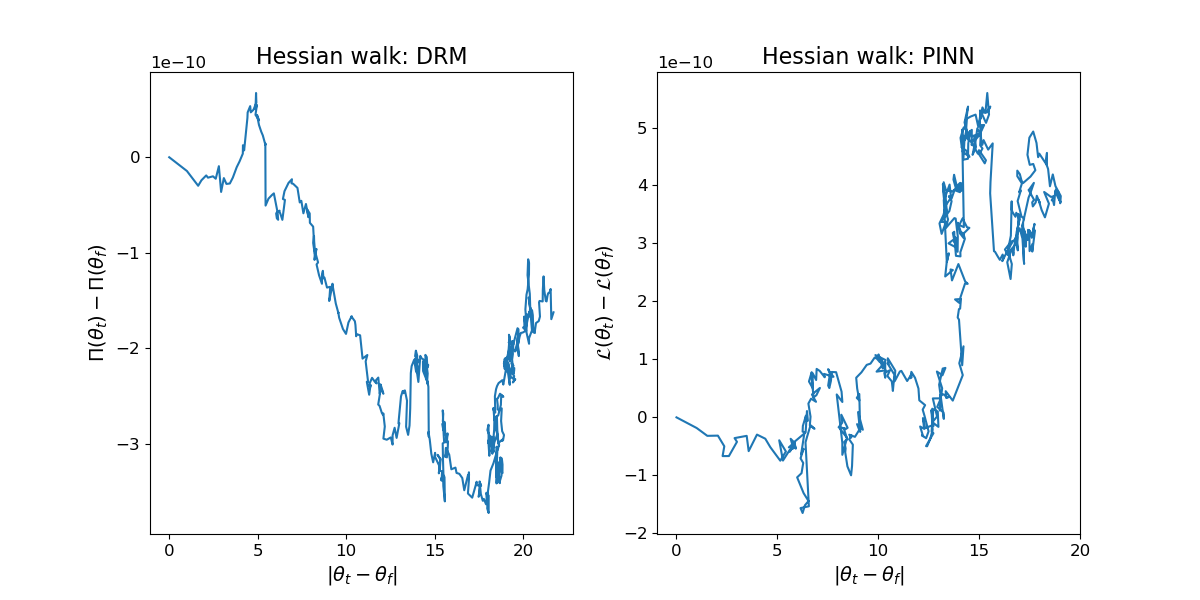}
\caption{Compared to the norm of the initial parameters $\bs \theta_f$, the Hessian walk travels appreciable distances in parameter space without leaving the isocontour of the two loss functions.}
\label{hessian_walk_fig}
\end{figure}

The eigenvector $\mbf v_{\text{min}}$ corresponds to a minimum curvature direction in parameter space. Given that the walk begins at a stationary point, minimum curvature directions correspond to directions in which small changes to the parameters do not appreciably change the loss. We can thus attempt to explore the space of parameters corresponding to equal values of the loss using the Hessian walk. In practice, we find that it is not necessary to set the step size to be small in order to remain on an isocontour of the loss. Using automatic differentiation to compute the Hessian of each objective function at each step $t$, we perform the walk of Eq. \eqref{hessian_walk} with a step size of $\eta=1$ for $500$ steps. See Figure \ref{hessian_walk_fig} for the results, on which we plot the change in the loss from its converged value as a function of the distance between the current parameters $\bs \theta_t$ and the starting point of the Hessian walk $\bs \theta_f$. In both the case of DRM and PINNs, the loss changes by less than $10^{-9}$ over the course of the $500$ steps. Note that the maximum distance the walk travels from the initial parameters is approximately $|\bs \theta_t - \bs \theta_f|=20$. The norm of the converged parameters from the optimization problem is $|\bs \theta_f|\approx 8$ for both DRM and PINNs, indicating that the walk has traveled $250\%$ of the norm of the converged parameters without changing the value of the loss. We will show in a subsequent subsection that, in many cases, the Hessian walk can be continued indefinitely, maintaining the exact loss value at arbitrary distances from the initial parameters.

\subsection{Mode connectivity}

\paragraph{} A number of works have noted that two distinct solutions obtained from different initializations and/or optimizers are separated by a high loss barrier with a linear connection, but a path can be found between them over which the loss never increases \cite{draxler_essentially_2019, garipov_loss_2018, fort_large_2019}. This phenomenon has been named ``mode connectivity.'' In this example, we test whether this phenomenon also appears in physics-informed problems. We randomly initialize the DRM and PINN networks two times and train each to a solution. In each case, we denote the two different solution parameters $\bs \theta_1$ and $\bs \theta_2$. We linearly connect the two solutions with the path $\bs \theta_1 +t(\bs \theta_2 - \bs \theta_1)$ for $t\in[0,1]$. Then, we use a quadratic Bezier curve with a control point $\mbf p$ to construct the curved path between the two solutions. This path is given by $(1-t)^2\bs \theta_1 +2t(1-t)\mbf p + t^2 \bs \theta_2$ for $t \in [0,1]$. We then solve an optimization problem for the position of the control point:

\begin{equation}\label{mode_connectivity}
     \underset{\mbf p}{\text{argmin }} \frac{1}{2}\int_0^1\Big( \Pi((1-t)^2\bs \theta_1 + 2t(1-t)\mbf p + t^2 \bs \theta_2) - \Pi(\bs \theta_1) \Big)^2 dt.
\end{equation}

\begin{figure}[hbt!]
\centering
\includegraphics[width=0.99\textwidth]{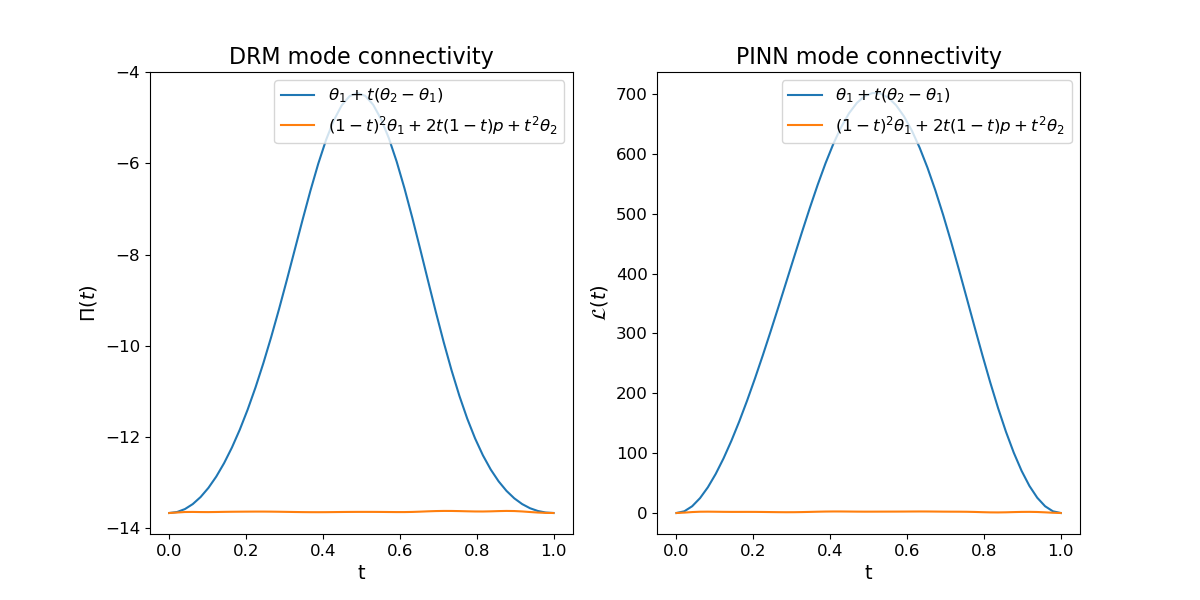}
\caption{For both DRM and PINN, the two solutions are separated by a high loss barrier with a linear connection, but a quadratic Bezier path can be found over which the loss does not increase. This shows that though the optimizer does not find the same solution for different initializations, the two solutions lie in the same basin.}
\label{1d_connectivity}
\end{figure}

This loss function penalizes departures from the initial loss value $\Pi(\bs \theta_1)$. A similar optimization problem is constructed with the PINN loss $\mathcal{L}$. If the modes are connected, we expect the value of Eq. \eqref{mode_connectivity} at the optimal control point to be approximately zero. See Figure \ref{1d_connectivity} for the results of the linear and Bezier curve connections. For both objectives, a high loss barrier separates the solution along the linear path, but the curved path is flat. Similar to the previous subsection, and in line with the findings from the loss landscape literature, mode connectivity suggests that there are equal loss ``valleys'' or ``tunnels'' connecting solutions. Whereas the Hessian walk showed that it was possible to venture far away from the converged solution without increasing the loss, this mode connectivity result reveals a particular structure in the landscape, which is that different initializations give rise to solutions which lie in the same basin of the loss landscape.

\subsection{Random directions}

\begin{figure}[hbt!]
\centering
\includegraphics[width=0.99\textwidth]{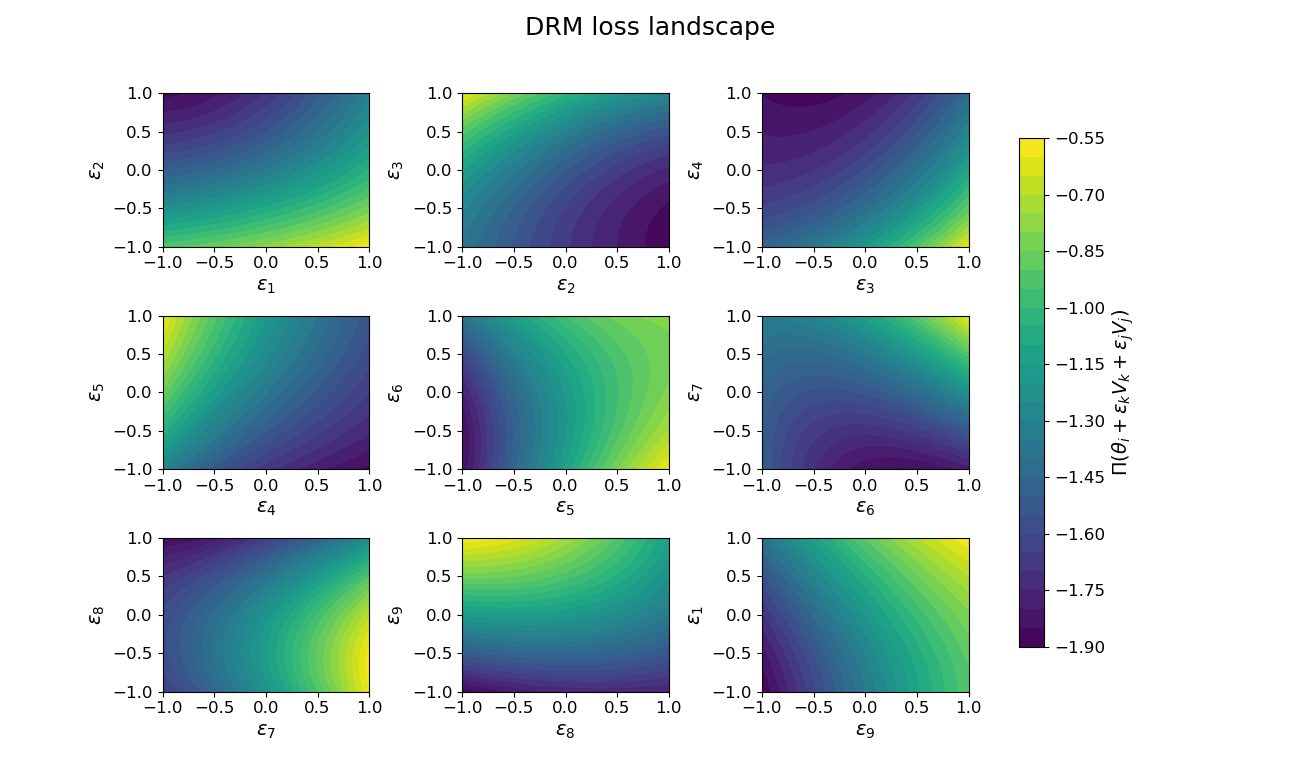}
\caption{DRM loss landscape around randomly initialized parameters $\bs \theta_i$ in $9$ combinations of random directions.}
\label{drm_init_random}
\end{figure}

\begin{figure}[hbt!]
\centering
\includegraphics[width=0.99\textwidth]{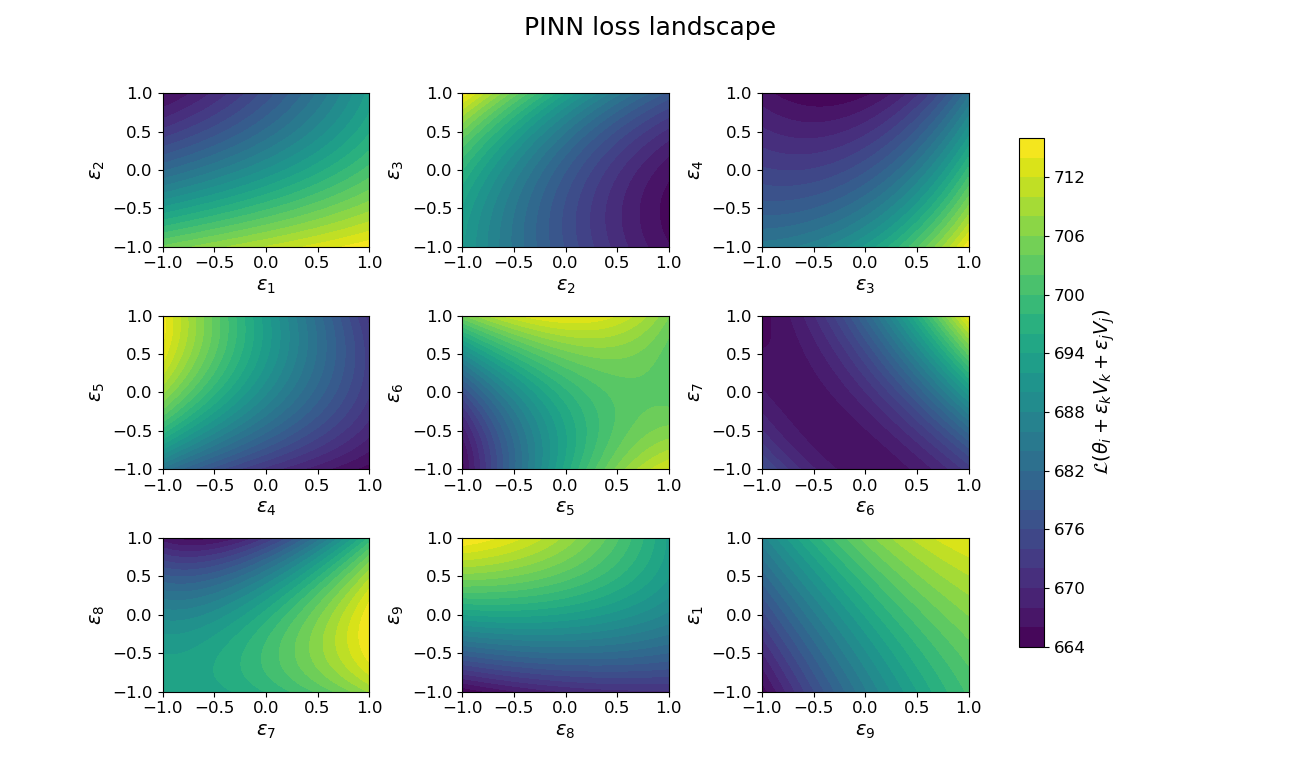}
\caption{PINN loss landscape around the same randomly initialized parameters as in DRM in $9$ combinations of the same random directions. A saddle in the fifth plot demonstrates the non-convexity of the loss landscape, though the loss surface is neither noisy nor ill-conditioned.}
\label{pinn_init_random}
\end{figure}

\begin{figure}[hbt!]
\centering
\includegraphics[width=0.99\textwidth]{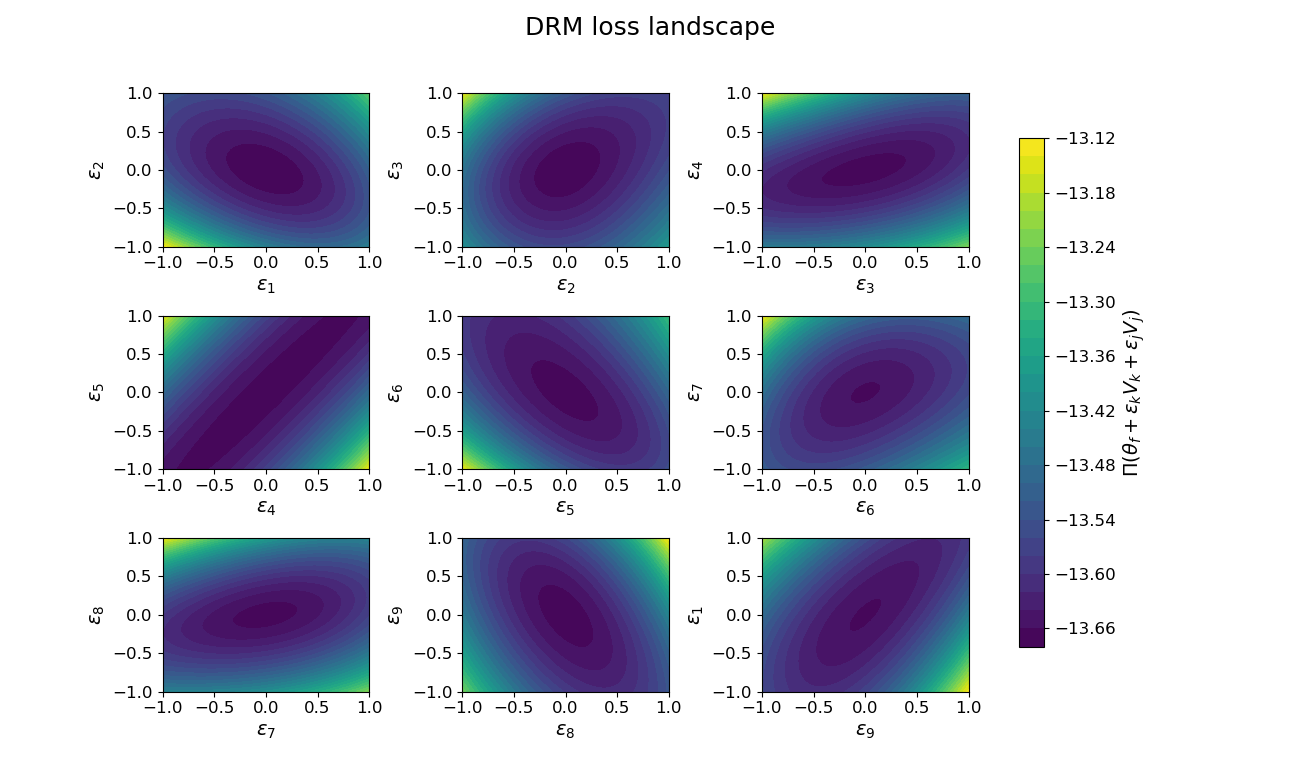}
\caption{DRM loss landscape around the solution parameters $\bs \theta_f^{\text{DRM}}$ shown in $9$ combinations of random directions. The loss is convex and well-conditioned in the vicinity of the solution.}
\label{drm_final_random}
\end{figure}

\begin{figure}[hbt!]
\centering
\includegraphics[width=0.99\textwidth]{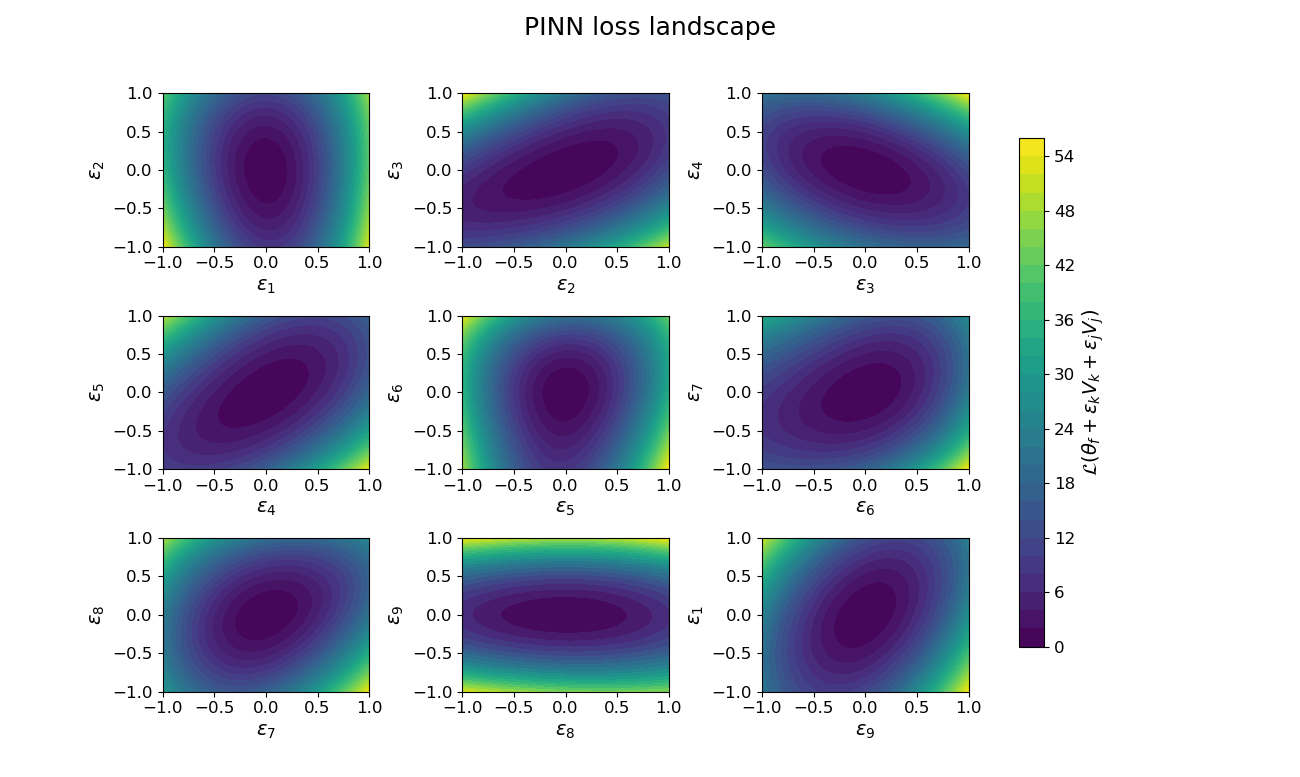}
\caption{PINN loss landscape around the solution parameters $\bs \theta_f^{\text{PINN}}$ shown in $9$ combinations of random directions. The loss is convex and well-conditioned in the vicinity of the solution.}
\label{pinn_final_random}
\end{figure}

\paragraph{} A straightforward strategy to visualize the loss landscape is to view cross-sections. Cross-sectional surface plots power the intriguing visualizations provided in \cite{ideami_loss_2026}. Given that loss surfaces are high-dimensional, there are an enormous number of combinations of directions over which to plot the loss surface, as well as a continuum of choices for the point about which the plot is made. The approach taken in \cite{li_visualizing_2018} is to plot the loss surface over planes defined by random directions in parameter space. This is the method we adopt here. For the DRM and PINN objectives, we generate contour plots with coordinates $\varepsilon_k, \varepsilon_j \in [-1,1]$ with:

\begin{equation*}
    \Pi( \tilde{\bs \theta}^{\text{DRM}} + \varepsilon_k \mbf V_k + \varepsilon_j \mbf V_j), \quad \mathcal L( \tilde{\bs \theta}^{\text{PINN}} + \varepsilon_k \mbf V_k + \varepsilon_j \mbf V_j), \quad k,j=1,2,\dots, \quad k\neq j,
\end{equation*}

\noindent where $ \tilde{\bs \theta}^{\text{DRM}}$ and $ \tilde{\bs \theta}^{\text{PINN}}$ define the centers of the contour plots and the $\mbf V_1,\mbf V_2,\dots$ are random unit vectors in $\mathbb R^{|\bs \theta|}$. We generate contour plots for $9$ combinations of random directions for both the DRM and PINNs. Using the two hidden-layer network of width $20$, Figures \ref{drm_init_random} and \ref{pinn_init_random} show these random cross-sections around a random initialization $\tilde{\bs \theta}^{\text{DRM}}=\tilde{\bs \theta}^{\text{PINN}}=\bs \theta_i$. We then obtain a solution to the DRM and PINN problems, providing converged parameters $\bs \theta^{\text{DRM}}_f$ and $\bs \theta^{\text{PINN}}_f$, which we use to center additional contour plots for both objectives given in Figures \ref{drm_final_random} and \ref{pinn_final_random}. To clarify, the parameters which solve the two problems are not equal, thus these two contour plots are not centered at the same point. We remark that the DRM and PINN loss landscapes at initialization show the same qualitative features in each of the $9$ cross-sections. For both objectives, the loss is convex in the vicinity of the solution in these random planes. The loss surfaces are also smooth and well-conditioned. We remark that the same qualitative behavior of the loss in the random planes is observed for MLP networks of the same width with depth $3$, $4$, and $5$. Around the solution, the loss appears convex, whereas around the random initialization, non-convexity is observed through the presence of saddle points.

\subsection{Random directions with stochastic objective}

\begin{figure}[hbt!]
\centering
\includegraphics[width=0.99\textwidth]{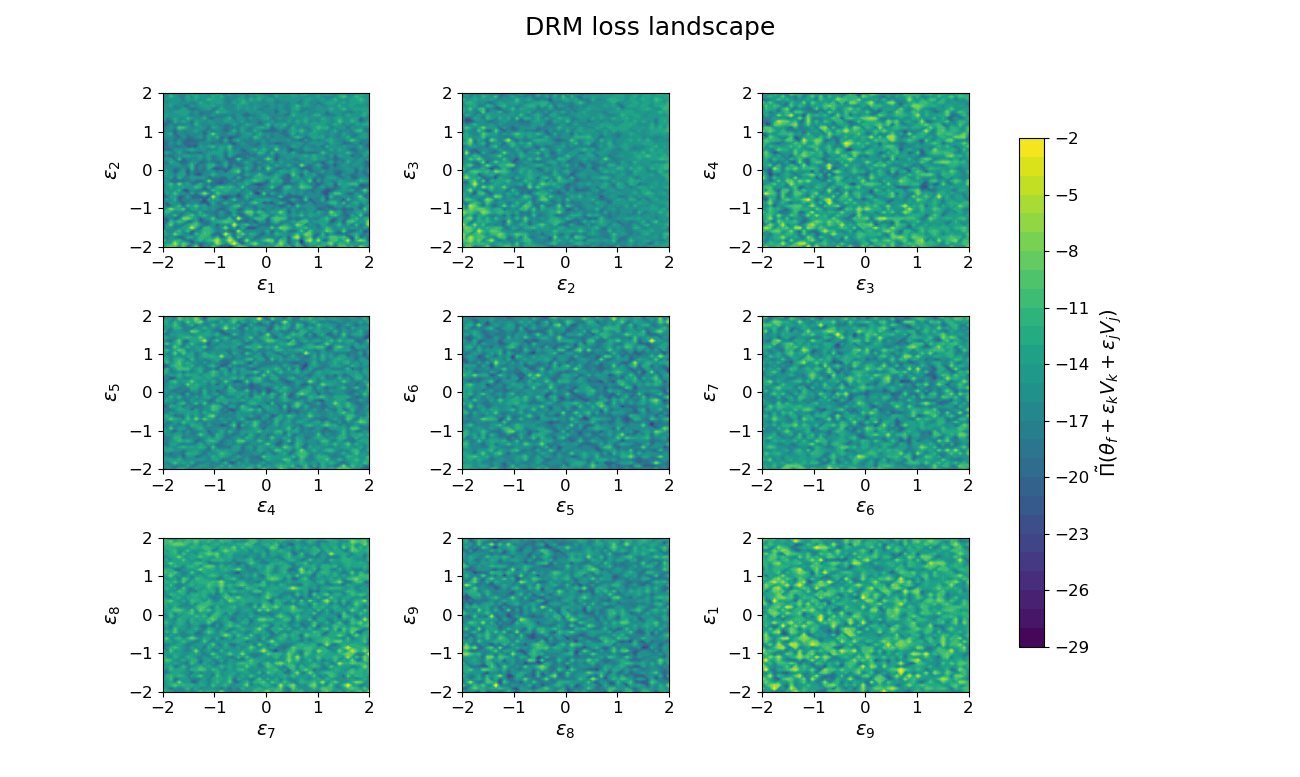}
\caption{Randomly integrating the DRM loss function destroys the structure in the loss landscape, giving rise to loss surfaces which resemble pure noise. In light of the large variance of integrand in DRM, the Monte Carlo integration is expected to be noisy.}
\label{stochastic_drm}
\end{figure}

\begin{figure}[hbt!]
\centering
\includegraphics[width=0.99\textwidth]{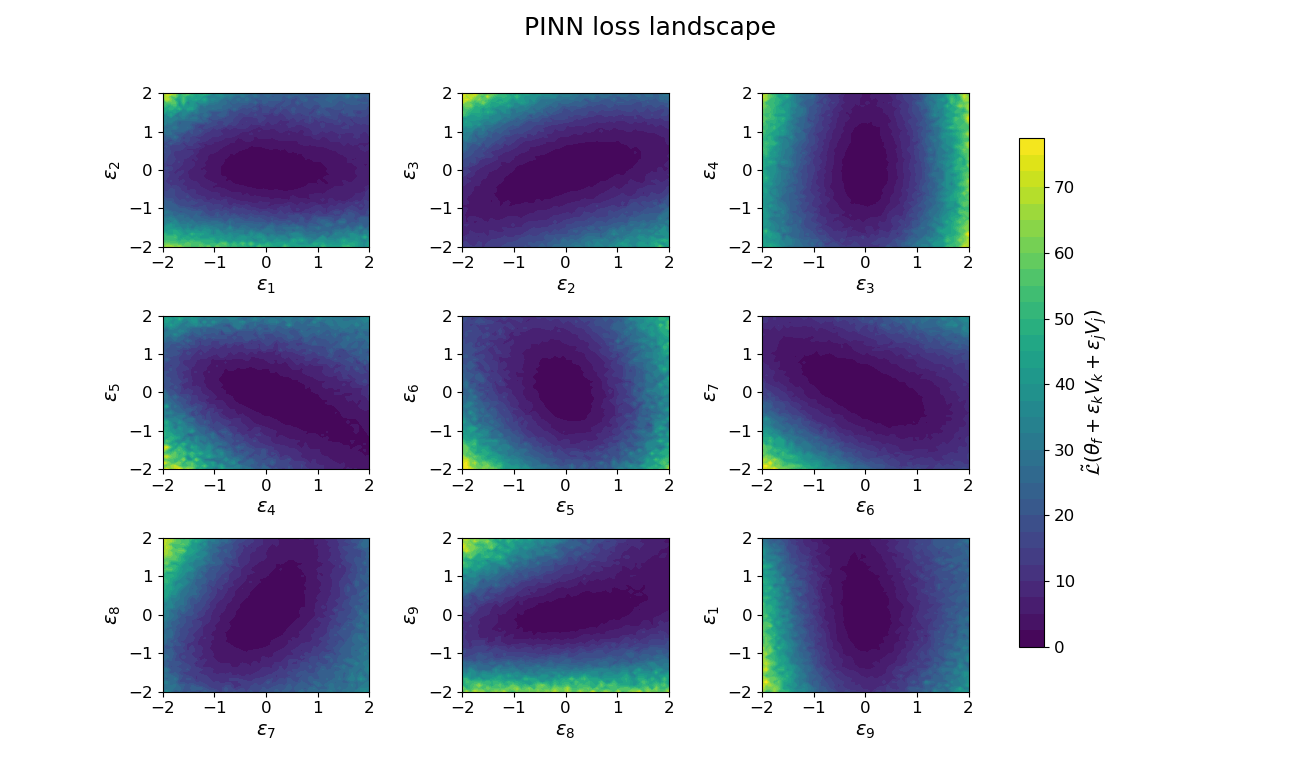}
\caption{Randomly integrating the PINN loss function leaves the general convex structure of the loss surface around the solution intact. However, the loss surfaces becomes rough, indicating noisy gradients.}
\label{stochastic_pinn}
\end{figure}

\paragraph{} Physics-informed training is often conducted with randomly sampled collocation points. This is analogous to mini-batching the data in the classification problems typically studied in the loss landscape literature. Until this point, the integration of the DRM and PINN objectives has been carried out with deterministic quadrature routines, namely mid-point integration on a fixed grid. In this example, we plot the loss landscapes defined by stochastic integration routines. The loss is computed with Monte Carlo integration every time it is queried. As such, the two loss functions become:

\begin{equation*}
    \tilde \Pi(\bs \theta) = \frac{1}{B} \sum_{i=1}^B\qty[ \frac{1}{2}\qty( \pd{u(x_i;\bs\theta)}{x})^2 - f(x_i) u(x_i; \bs \theta)], \quad \tilde{\mathcal L}(\bs \theta) = \frac{1}{2B} \sum_{i=1}^B\qty[ \pdd{u(x_i;\bs \theta)}{x} + f(x_i) ]^2 ,
\end{equation*}

\noindent where the $x_i \overset{\text{iid}}{\sim} \mathcal{U}(0,1) $. Like the previous subsection, we plot these two objective functions in random planes. At each parameter setting used to generate the contour plots, the Monte Carlo integration grid is re-sampled. This is meant to provide insight into the loss landscape suggested by gradients computed from a stochastic optimizer. See Figure \ref{stochastic_drm} for the stochastic DRM loss landscape and \ref{stochastic_pinn} for the stochastic PINN loss landscape in random directions about the converged solution. Note that we do not use random integration to obtain the solution parameters about which the plots are centered. These figures are generated with an integration batch size of $B=100$. Surprisingly, the random integration destroys the structure of the Deep Ritz loss, whereas the convex contours remain intact for the PINN loss landscape, albeit with the addition of some noise. To the best of our knowledge, this dramatic difference between the random integration of PINN and DRM loss functions has not been noted in the literature. This is a consequence of the large variance of the integrand for DRM. The variance of the Monte Carlo estimator of the integral is $\sigma^2/B$, where the variance $\sigma^2$ for DRM is

\begin{equation*}
    \sigma^2 = \text{Var}\qty( \frac{1}{2}\qty(\pd{u}{x})^2-fu) = \int_0^1 \qty( \frac{1}{2} \qty(\pd{u}{x})^2 - fu)^2 dx - \qty(\int_0^1 \frac{1}{2}\qty(\pd{u}{x})^2 - fudx)^2.
\end{equation*}

At the solution to our chosen problem, the variance from DRM is $5$ orders of magnitude larger than the variance of the PINNs integrand. This explains why the DRM loss landscape has the appearance of pure noise, whereas the PINN loss retains its structure.

\subsection{Hessian eigenvalues}

\begin{figure}[hbt!]
\centering
\includegraphics[width=0.99\textwidth]{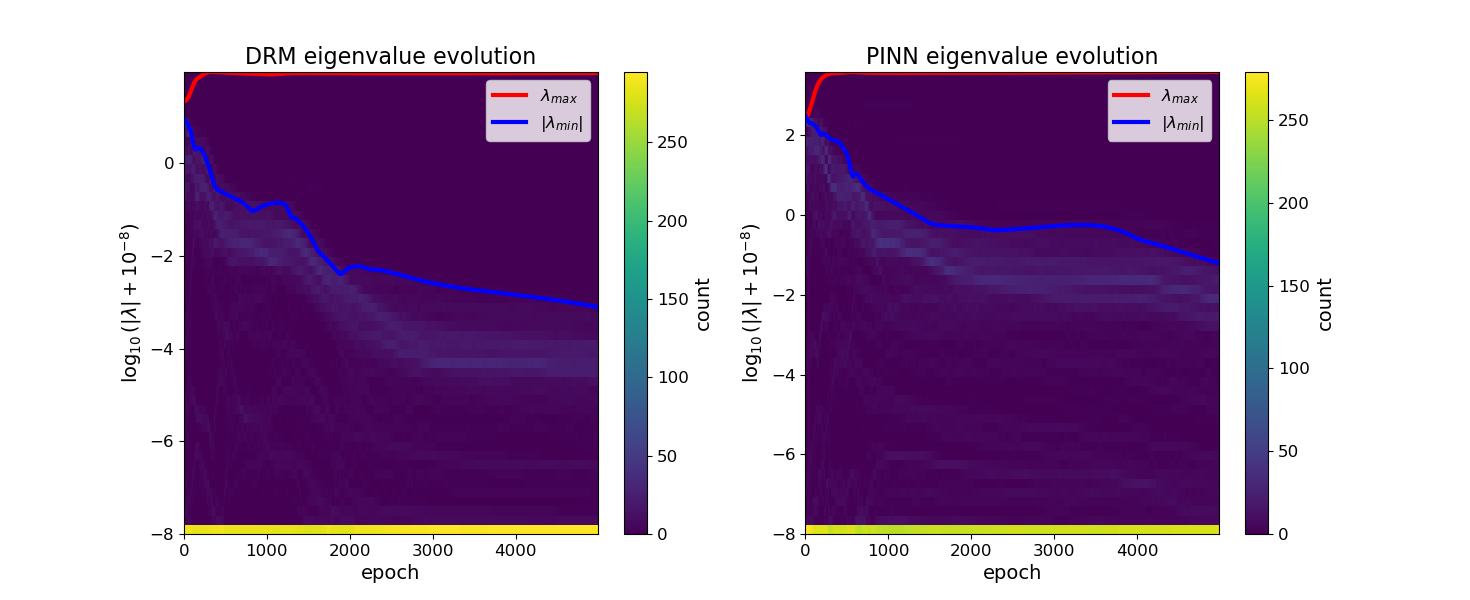}
\caption{As observed in the existing loss landscape literature, the majority of Hessian eigenvalues concentrate around zero. The faint band in the plots indicates that a set of moderately sized eigenvalues decay to zero over the course of optimization. The maximum eigenvalue increases, whereas the negative curvature disappears as the optimizer approaches a solution.}
\label{1d_evolution}
\end{figure}

\paragraph{} The eigenvalues of the Hessian matrix of the objective provide curvature information about the loss landscape. Whereas we used random directions in the previous subsection, the Hessian eigenvalues have been used to generate contour plots in directions of maximum curvature \cite{bottcher_visualizing_2024}. The spectrum of the Hessian eigenvalues is also a concise characterization of the local geometry of the loss landscape \cite{sagun_eigenvalues_2017, sagun_empirical_2018}. Here, we attempt to characterize the landscape that the optimizer moves through by plotting the evolution of the eigenvalue spectrum over the course of training for both the DRM and PINN objectives. To do this, we make a two-dimensional histogram of the eigenvalue magnitude as the vertical axis and the optimization epoch on the horizontal. Note that we take the absolute value of the eigenvalues so that a log scale can be used, though this obscures the negative curvatures observed in the eigenvalue spectrum. We also add a small constant shift to the eigenvalue magnitude in order to avoid the plot spanning many orders of magnitude which are all indistinguishable from zero. See Figure \ref{1d_evolution} for the results. Over the course of the optimization process, the overwhelming majority of eigenvalues for both objectives are zero. The histogram shows a faint band of moderately sized eigenvalues which decay toward zero over the course of optimization. However, the maximum eigenvalue monotonically increases over the course of optimization. We also plot the evolution of the magnitude of the most negative eigenvalue $\lambda_{\text{min}}$, which starts off on the order of $10^1$-$10^2$ and decays to approximately zero. These plots suggest that the optimizer finds a basin which is steep-sided with respect to a small number of eigenvector directions, and is flat in the remaining directions. This flatness suggests that the spatial representation built by the network is redundant, as many directions in parameter space leave the solution, and thus the loss, unchanged. We note that the Hessian singularity is not a function of saturation of the hyperbolic tangent activation functions. A custom activation function which does not saturate in either direction can be constructed and used in the neural network representation. An example of such an activation is $x + \tanh x$. We verify that even with this activation function, the spectrum of the Hessian still concentrates around zero.

\paragraph{} To understand one source of the singularity of the Hessian, we write the neural network discretization of the state field as

\begin{equation*}
    u(x; \bs \theta) = \sum_{i=1}^N \theta_i^O \sin( \pi x) \tilde h_i(x; \bs \theta^I),
\end{equation*}

\noindent where the total set of parameters comprises an ``outer'' and ``inner set,'' i.e., $\bs \theta=[\bs \theta^O , \bs \theta^I]$. We think of the inner set as building basis functions $\{ h_i(x) \}_{i=1}^N$, where $h_i(x) = \sin(\pi x) \tilde h_i(x )$. This shows that the distance functions used to enforce the homogeneous Dirichlet boundary conditions are incorporated at the final layer, as opposed to multiplying the network output. This allows the final layer to be interpreted as basis functions which satisfy the boundary conditions, much like standard finite element or spectral approaches to solving PDEs. The outer parameters are then coefficients on the basis functions. Thus, an MLP network can be thought of as a discretization strategy in which both the basis functions and their coefficients are learnable. Now, if we can find a direction $\mbf v$ for which the state field is constant, changing the parameters in this direction will not change the loss value. Thus, this direction will be in the span of the Hessian eigenvectors with zero eigenvalues. Suppose that the current parameter setting is the converged parameter set $\bs \theta^f$. The basis functions are then $\{ h_i(x;\bs \theta^{f,I})\}_{i=1}^N$. We populate a matrix with the current basis functions as follows:

\begin{equation*}
    \mbf H(x) = \begin{bmatrix}
         h_1(x; \bs \theta^{f,I}) &  h_2(x; \bs \theta^{f,I}) & \dots &  h_N(x; \bs \theta^{f,I})
    \end{bmatrix}.
\end{equation*}

Note that in practice, we discretize the basis functions on the integration grid, such that $\mbf H \in \mathbb R^{100 \times N}$, where we remind the reader that $100$ is the number of integration points. Now, if any of these basis functions are linearly dependent, the matrix $\mbf H(x)$ has a nullspace, meaning that there is some change to the outer parameters $\bs \Delta$ such that 

\begin{equation*}
    \mbf H(x) ( \bs \theta^{O,f} + \bs \Delta ) = \mbf H(x)  \bs \theta^{O,f} = u( x; \bs \theta^f).
\end{equation*}

\begin{figure}[hbt!]
\centering
\includegraphics[width=0.99\textwidth]{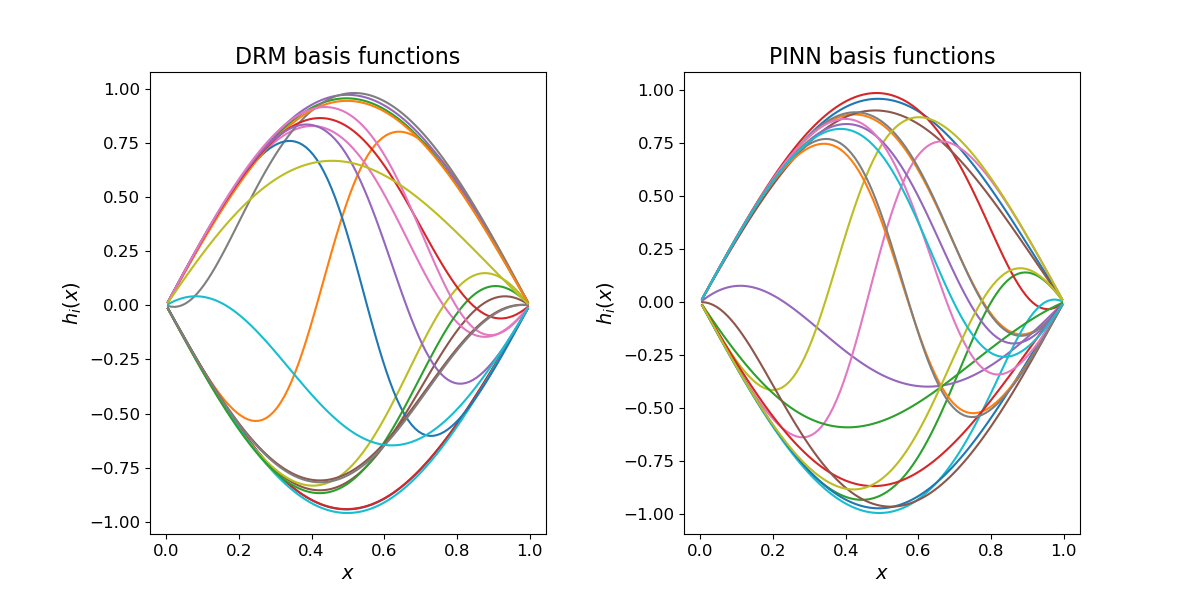}
\caption{For both DRM and PINNs, the learned basis functions are not independent, meaning that there are coordinated changes to their coefficients which leave the state field unchanged. The linear dependence of the basis is one source of Hessian singularity.}
\label{1d_basis}
\end{figure}

In the case of a dependent basis, the coefficients can be adjusted by $\bs \Delta$ without altering the state field. Thus, one possible source of Hessian singularity is linearly dependent basis functions, as the vector $\mbf v=[\bs \Delta , \mbf 0]$ keeps the basis functions fixed and exploits the nullspace of the basis function matrix to hold the state field constant. Whether the network learns independent basis functions is an empirical question. Training the networks to convergence with ADAM optimization for both DRM and PINN objectives, we obtain basis functions as shown in Figure \ref{1d_basis}. Visually, there appears to be significant redundancy in the $N=20$ basis functions. To verify this, we compute the Gram matrix $G_{ij} = \int_0^1 h_i(x) h_j(x) dx$ and determine its rank by counting the number of eigenvalues greater than $1 \times 10^{-12}$ of the maximum eigenvalue. For DRM, the rank is $11$ and for PINNs, the rank is $13$. In both cases, the basis functions are effectively dependent, which explains some (but not all) of the flat directions in the loss landscape around the solution. We remark that the linear dependence of the basis functions suggests the existence of arbitrarily long straight-line isocontours of the loss function. In other words, we have for the DRM loss that

\begin{equation*}
    \Pi( \bs \theta^f ) = \Pi\qty( \bs \theta^f + t \begin{bmatrix}
        \bs \Delta \\ \mbf 0
    \end{bmatrix} ) \quad \forall t.
\end{equation*}

\begin{figure}[hbt!]
\centering
\includegraphics[width=0.99\textwidth]{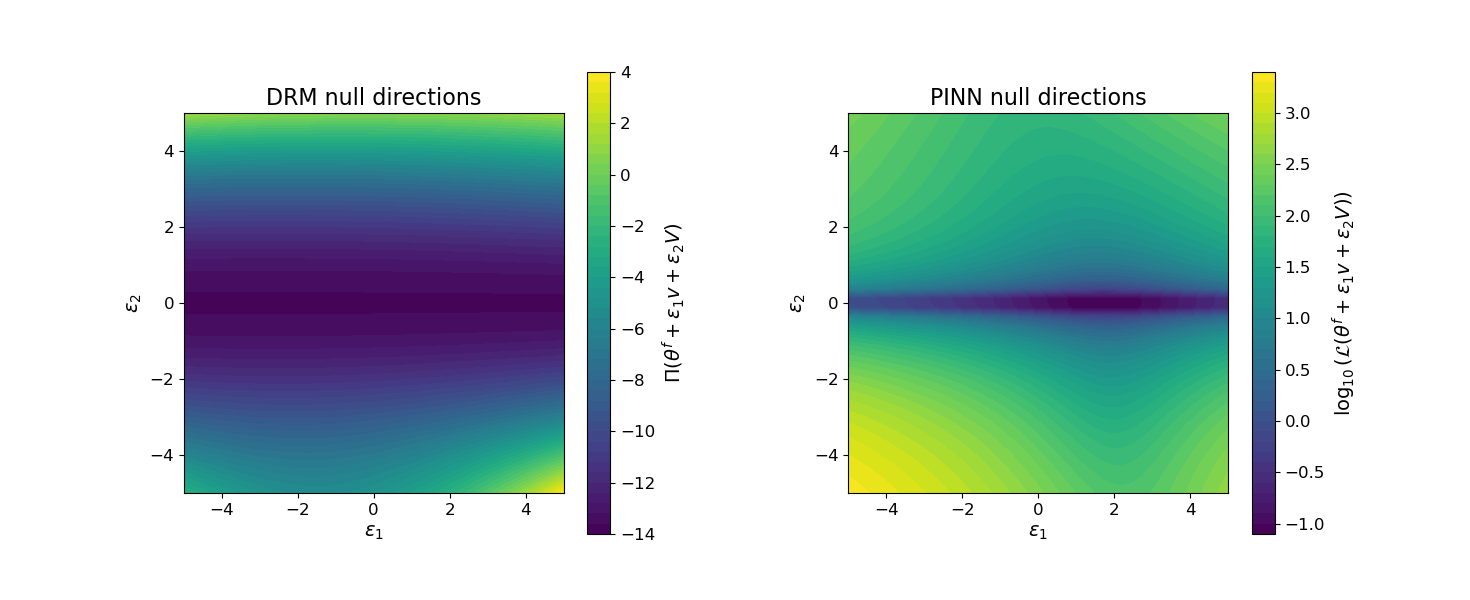}
\caption{The nullspace of the matrix of basis functions gives rise to flat directions in parameter space which extend arbitrarily far in both directions. We plot the logarithm of the PINN objective to better visualize this ``trench'' in the loss landscape.}
\label{1d_trench}
\end{figure}

Of course, the same can be said for the PINN loss. To visualize these null directions, we plot the two loss functions in a plane defined by the null direction $\mbf v = [\bs \Delta , \mbf 0]$, where $\bs \Delta$ is the eigenvector corresponding to the minimum eigenvalue of the Gram matrix, and a random direction $\mbf V$. In both cases, the contour plots are centered on the converged parameters obtained from the two loss functions. See Figure \ref{1d_trench} to visualize these null directions. The visualization is especially striking when plotting the logarithm of the PINN objective, which we avoid in the case of DRM because of the negative energy values. These straight-line isocontours of the loss extend arbitrarily in both directions of parameter space. We remark that, by adjusting the nullspace of the matrix of basis functions, changes to the inner parameters change these null directions. Though so long as the basis functions are dependent, there will always be null directions of this sort.

\subsection{Goldilocks zone}

\begin{figure}[hbt!]
\centering
\includegraphics[width=0.99\textwidth]{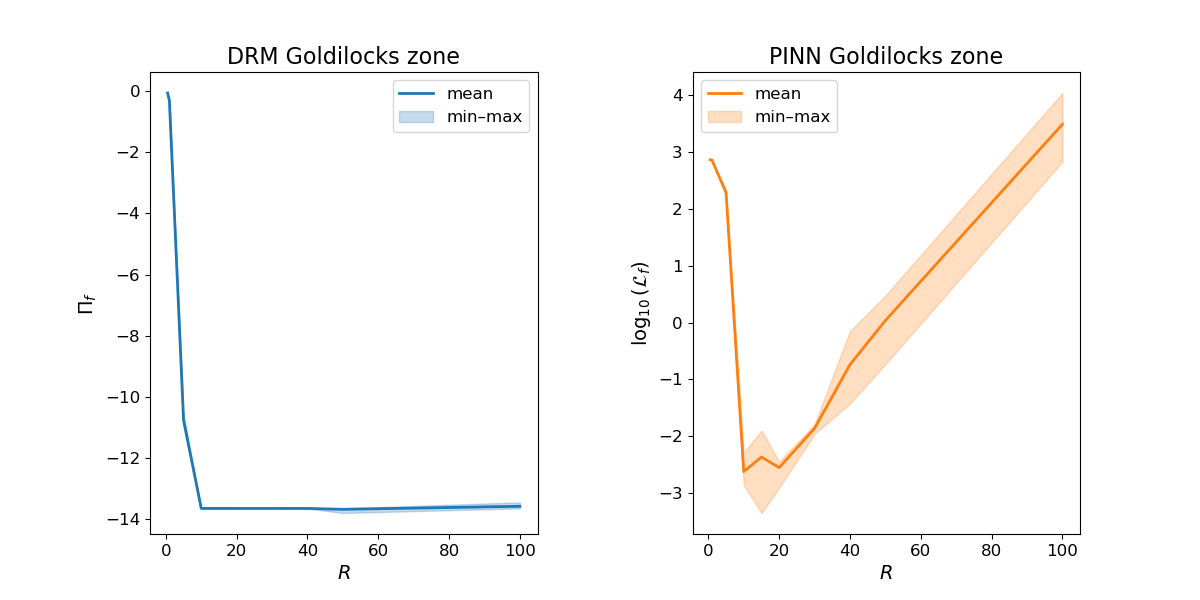}
\caption{Both DRM and PINN training problems fail when the parameters are confined to a hypersphere with too small of a radius. In the case of DRM, large hypersphere constraints do not significantly impact performance, whereas the converged PINN objective increases monotonically with the radius after passing through the ``Goldilocks zone.'' }
\label{1d_goldilocks}
\end{figure}

\begin{figure}[hbt!]
\centering
\includegraphics[width=0.99\textwidth]{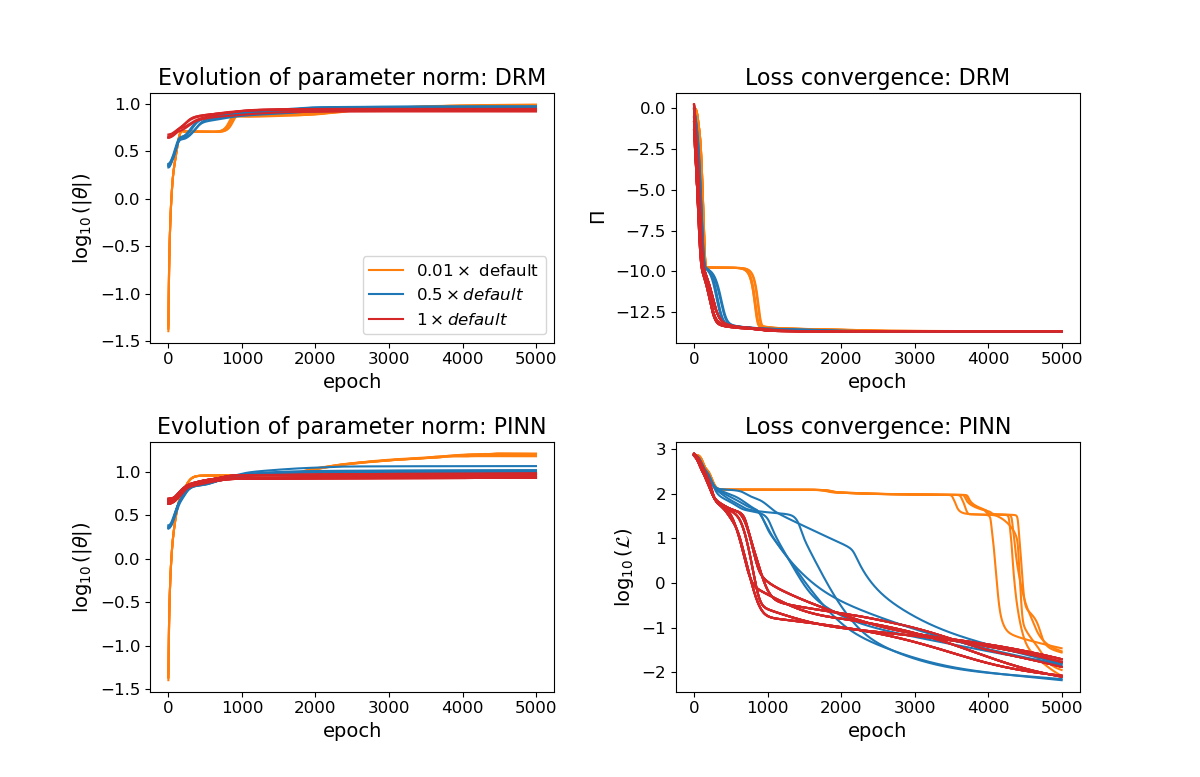}
\caption{With both DRM and PINNs, parameters initialized with small radius increase until entering the Goldilocks zone observed in our constrained optimization experiment. The Goldilocks radius corresponds closely with magnitude of the parameters from PyTorch's default initialization.}
\label{1d_radius}
\end{figure}

\paragraph{} The authors in \cite{fort_goldilocks_2018} find that the loss landscape is especially convex in certain radii in parameter space, leading to faster and more successful training. We explore whether this result for holds for physics-informed networks as well. In particular, for both DRM and PINNs, we restrict training to the surface of hyperspheres of varying radii, and compare the final loss values obtained after a fixed number of epochs. The following algorithm is used to enforce the hypersphere constraint on the parameter trajectories:

\begin{equation*}
    \tilde{\bs \theta}_{t+1} = \bs \theta_t - \eta \mathcal U(\bs \theta_t), \quad \bs \theta_{t+1}= R\frac{\tilde{\bs \theta}_{t+1}}{\lVert  \tilde{\bs \theta}_{t+1}\rVert},
\end{equation*}

\noindent where $\mathcal U$ denotes the step from the ADAM optimizer with either the DRM or PINN loss, $\eta$ is the learning rate, and $R$ is the radius of the given hypersphere. We report the final losses from each method as $\Pi_f$ and $\mathcal L_f$ respectively. With the two hidden-layer network of width $20$, we run ADAM optimization with a learning rate of $1 \times 10^{-3}$ for $5000$ epochs on $10$ different hypersphere radii. We perform $5$ trials at each radius value, and report the final values of the loss function. See Figure \ref{1d_goldilocks} for the results. For both DRM and PINNs, the network cannot find a solution on the small radii hypersphere. Interestingly, DRM has consistent performance for all radii $R\geq 10$, whereas PINNs exhibit the Goldilocks zone behavior, with significantly lower final loss values in the range $10 \leq R \leq 30$. We remark that this phenomenon demonstrates the importance of initialization, as initializing the parameters with the appropriate norm leads to solutions with lower loss values. However, we observe that when the restriction to the hypersphere surface is removed, parameters initialized with small norms move radially outwards to the so-called Goldilocks zone, which appears to be $R\approx 10$ for this problem. To illustrate this, we initialize the network with parameters at different radii and observe the evolution of the parameter radius over the course of training. We simply scale the parameters obtained by the default initialization by a factor of $0.01$ and $0.5$, then compare to the true default initialization. See Figure \ref{1d_radius} for the results. When the parameters are initialized at small radii, they move outward to the Goldilocks zone. For larger initializations, the parameter radius remains approximately constant over the course of training. However, we do observe that the larger radius initializations lead to worse performance on both the DRM and PINN objectives. As seen in Figure \ref{1d_goldilocks}, this difference is more significant in the case of the PINN objective.

\subsection{Intrinsic dimensionality}

\begin{figure}[hbt!]
\centering
\includegraphics[width=0.99\textwidth]{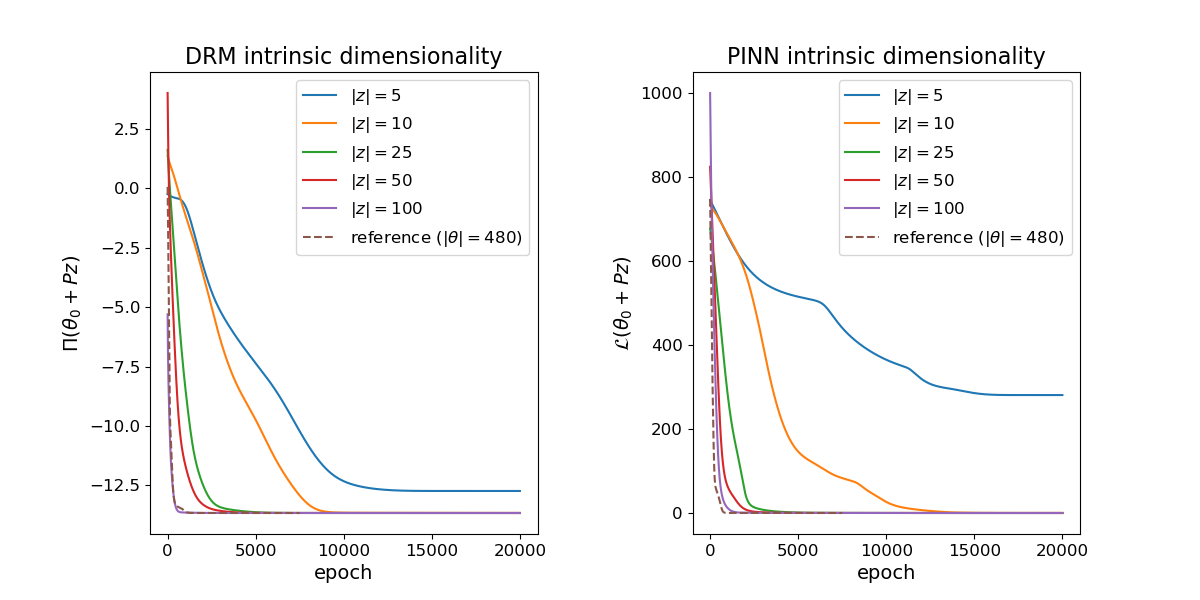}
\caption{When trained in random low-dimensional subspaces, both DRM and PINNs achieve the same converged loss value as the full network with a training hyperplane dimension of only $10$. The confinement to the subspace does delay convergence, however. }
\label{1d_intrinsic}
\end{figure}

\paragraph{} In \cite{li_measuring_2018}, the intrinsic dimensionality of the loss landscape is measured by training in a random affine subspace of the parameters. In the case of DRM and PINNs, the training problems become

\begin{equation*}
    \underset{\mbf z^{\text{DRM}}}{\text{argmin }} \Pi( \bs \theta^{\text{DRM}}_0 + \mbf P \mbf z^{\text{DRM}}), \quad \underset{\mbf z^{\text{PINN}}}{\text{argmin }} \mathcal L( \bs \theta^{\text{PINN}}_0 + \mbf P \mbf z^{\text{PINN}}),
\end{equation*}

\noindent where $\mbf P\in \mathbb R^{|\bs \theta| \times |\mbf z|}$ is a matrix whose columns are random unit vectors and $\mbf z$ are the coordinates in the low-dimensional subspace. We take the parameters defining the offset of the affine subspaces to be equal, i.e., $\bs \theta_0^{\text{DRM}}=\bs \theta_0^{\text{PINN}}$. Furthermore, this offset comes from PyTorch's default initialization. In this example, we sweep over $5$ subspace dimensions and train the network in the random subspace until convergence. We use the same two hidden-layer network with a width of $20$ and a learning rate of $1 \times 10^{-3}$. To ensure convergence, the training in the random subspace is run for $2 \times 10^4$ epochs. The random projection matrix $\mbf P$ is the same for both DRM and PINNs at each subspace dimension, but is re-initialized when the dimension changes. The initial parameters are recomputed with PyTorch default initialization at each new subspace dimension. See Figure \ref{1d_intrinsic} for the results. As expected, the rate of convergence increases with the dimension of the subspace. More surprising is that even low-dimensional networks obtain the same final loss value as the reference solution with $480$ parameters, which is trained without any low-dimensional embedding. Equivalent performance is obtained with just $10$ randomly chosen directions for both DRM and PINNs. 

\subsection{Optimization trajectory}

\begin{figure}[hbt!]
\centering
\includegraphics[width=0.99\textwidth]{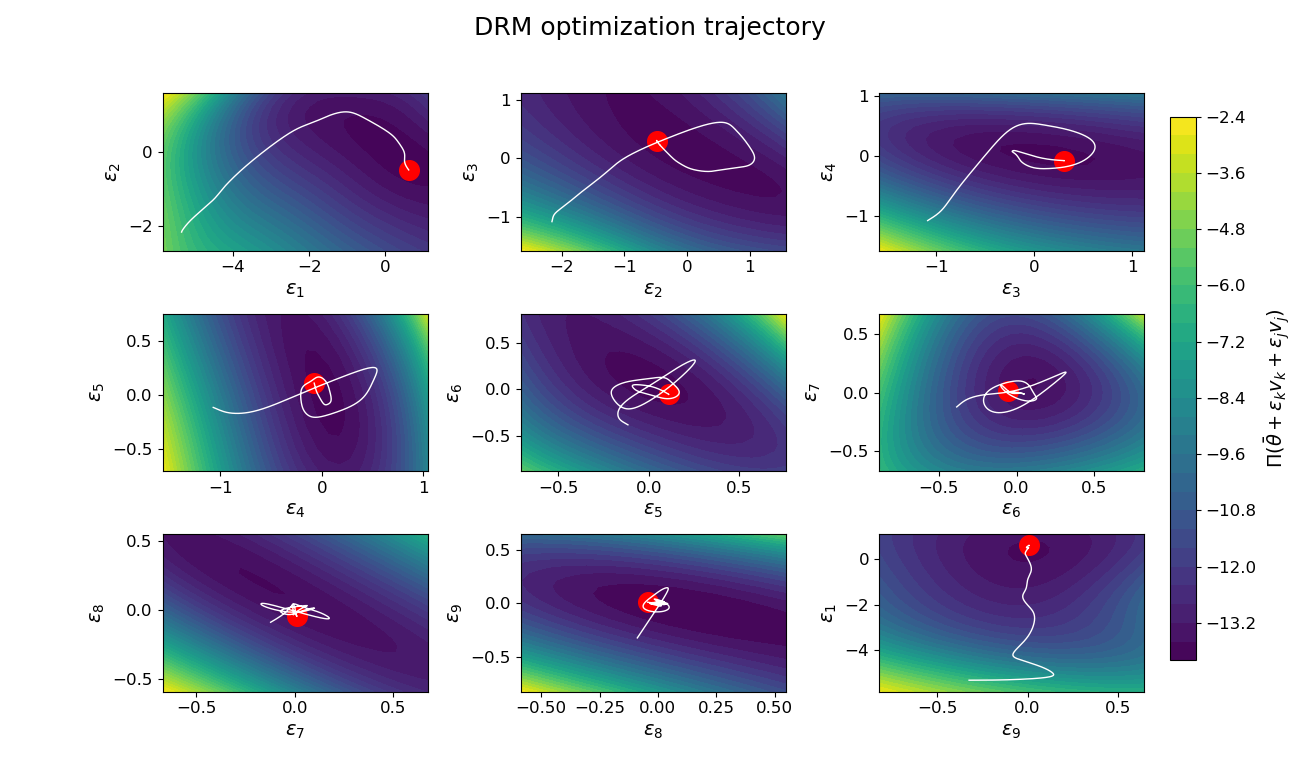}
\caption{The DRM loss is convex when viewed in a plane that explains the vast majority of the variance in the trajectory data. The planes spanned by higher-index principal components show training trajectories with more oscillations, though these planes capture only a small fraction of the variance. The projected final parameter value is indicated in red.}
\label{1d_drm_traj}
\end{figure}

\begin{figure}[hbt!]
\centering
\includegraphics[width=0.99\textwidth]{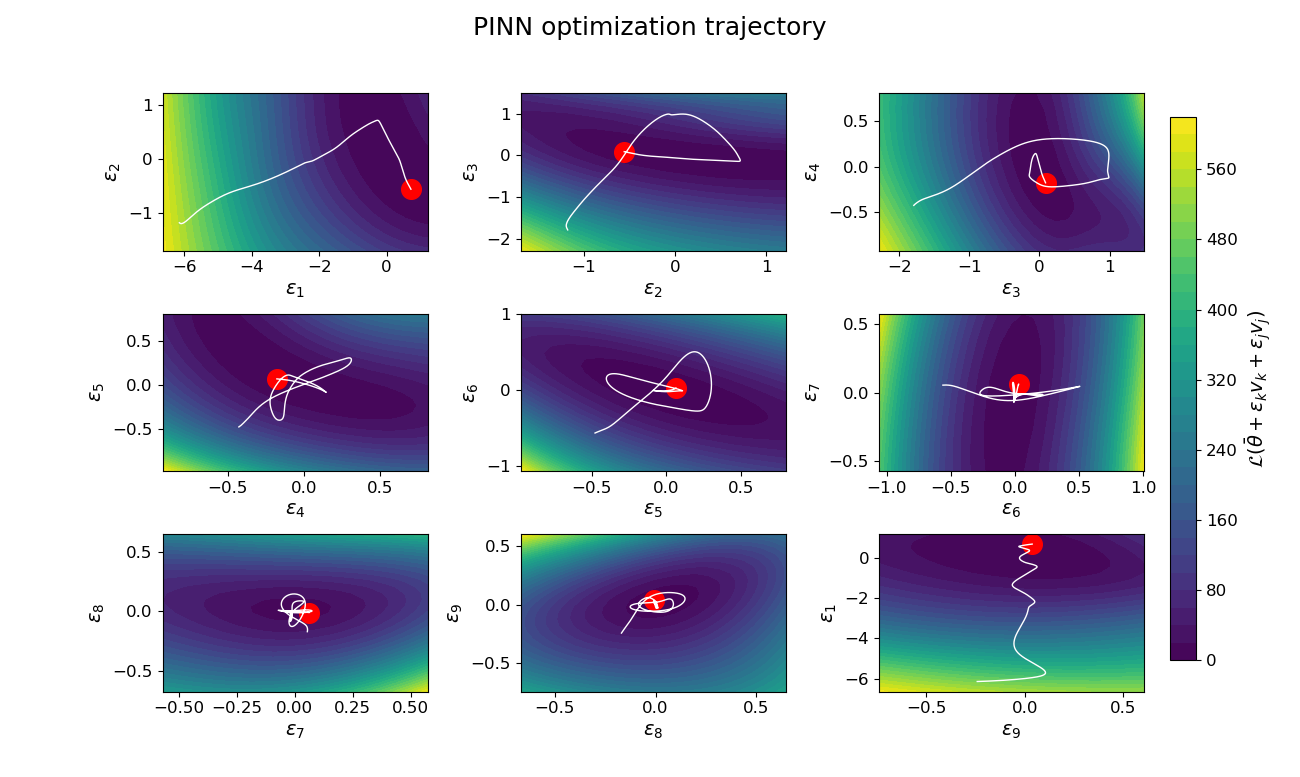}
\caption{Similar to DRM, the PINN loss is convex when viewed in the plane spanned by the first two principal components. We remark that both loss functions have the same qualitative features in the plane of maximum variance.}
\label{1d_pinn_traj}
\end{figure}

\paragraph{} Both \cite{li_visualizing_2018} and \cite{gur-ari_gradient_2018} find that the optimization trajectory of the neural network parameters lies in a low-dimensional subspace. Per \cite{li_visualizing_2018}, we plot the optimization dynamics of the parameters in this subspace. For both DRM and PINN training, we store the evolution of the parameters in a matrix given by

\begin{equation*}
    \bs \Theta = [ \bs \theta_0 , \bs \theta_1,\dots, \bs \theta_T],
\end{equation*}

\noindent where each column corresponds to the parameters at a given optimization epoch and $T$ is the total number of optimization epochs. We then perform principal component analysis (PCA) on the data matrix $\bs \Theta \in \mathbb R^{|\bs \theta| \times T}$ to find the directions of maximum variance. We call the principal components corresponding to decreasing variance $\mbf v_1,\mbf v_2,\dots,\mbf v_Q$, where we take only components up to some maximum order $Q$. We minimize the DRM and PINN objectives with ADAM optimization and the same width $20$ network, where both objectives are initialized at the same initial parameters $\bs \theta_i$. We find that in the case of DRM, the first and second principal components capture $78.7\%$ and $16.4\%$ of the variance of the trajectory data respectively. For the PINN training, the first and second principal components capture $79.7\%$ and $12.5\%$ of the variance. For the two objectives, the average parameter value, which we denote $\bar{\bs \theta}$, and two principal components $\mbf v_k$ and $\mbf v_j$ define a plane in parameter space. We denote the coordinates of this plane $\varepsilon_j$ and $\varepsilon_k$ where the origin is the average parameter value. Figures \ref{1d_drm_traj} and \ref{1d_pinn_traj} show the optimization trajectory projected onto planes spanned by combinations of the principal components. We remark that the optimization dynamics are simplest in the plane spanned by the first two principal components, in the sense that the path is closest to a straight line. In both cases, this first principal plane explains more than $90\%$ of the variance in the data, meaning that this projected trajectory gives a great deal of insight into the path followed by the optimizer. We remark that the loss is convex in this plane, and the MLI property thus holds. These plots suggest that for both DRM and PINNs, the parameters may not see parts of the loss landscape which suggest non-convexity.

\subsection{Acceleration}

\begin{figure}[hbt!]
\centering
\includegraphics[width=0.99\textwidth]{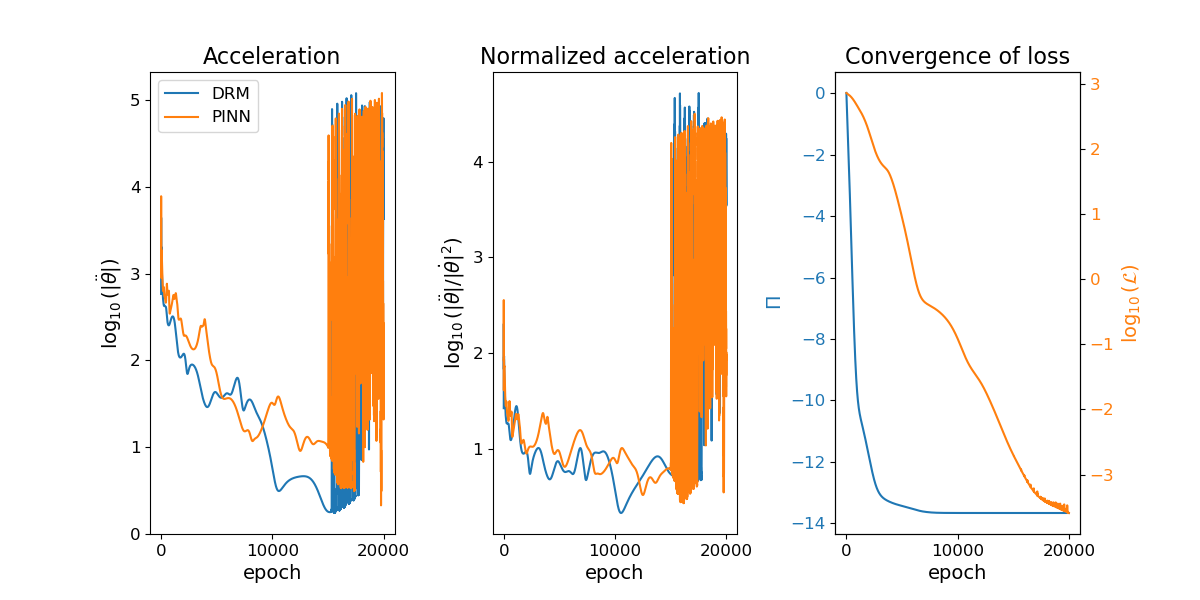}
\caption{The magnitude of the acceleration of the parameters decreases over the course of training, and then oscillates dramatically near the minimum. When the acceleration is normalized by the step size, the decreasing trend is less evident, indicating that the landscape is steeper in the initial phases of training.}
\label{1d_acceleration}
\end{figure}

\paragraph{} The MLI results suggest that the parameters could follow a straight-line path from initialization to the solution and monotonically decrease the loss. Of course, this is not to say that the parameters actually do this. It is natural to wonder how straight the trajectory of the parameters is. In the previous subsection, the trajectory projected into principal planes provided some insight into its curvature. Another technique to quantify straightness of the path is to measure the magnitude of the parameters' ``acceleration.'' Calling the learning rate $\eta$, we define the acceleration as

\begin{equation}\label{acceleration}
    \ddot {\bs \theta}_t = \frac{\bs \theta_{t+1} - 2 \bs \theta_t + \bs \theta_{t-1}}{\eta^2},
\end{equation}

\noindent which is simply a finite difference expression for the second derivative of the parameter vector in pseudo-time, where the training epoch plays the role of a discrete time variable. We propose the magnitude of the acceleration in Eq. \eqref{acceleration} as one measure of the straightness of the path in parameter space, as changes in the gradient are detected by this measure. That being said, the parameters can accelerate by the magnitude of the gradient changing as well. To account for this, we can normalize the acceleration by the squared norm of the velocity $|\dot{ \bs \theta}|^2$, where the velocity is finite differenced as $\dot{\bs \theta} = (\bs \theta_{t+1} - \bs \theta_t)/\eta$. This provides another measure of the curvature of the path. We solve the DRM and PINN problems with the same network architecture and ADAM optimization with a learning rate of $\eta=1\times 10^{-4}$. We run the problem for $2 \times 10^4$ epochs to illustrate post-convergence behavior of the parameter dynamics. See Figure \ref{1d_acceleration} for the results. For both DRM and PINNs, the magnitude of the acceleration decreases on the way to convergence, and then oscillates near the minima of the two objectives. When the acceleration is normalized by the step size, it is closer to a constant value in the pre-convergence regime. This suggests that the typical change in the direction of the update---which is not simply the gradient, given ADAM's use of momentum---is approximately constant over the course of the training, when accounting for the steepness of the loss landscape, as measured by velocity. 


\subsection{No bad local minima}

\begin{figure}[hbt!]
\centering
\includegraphics[width=0.99\textwidth]{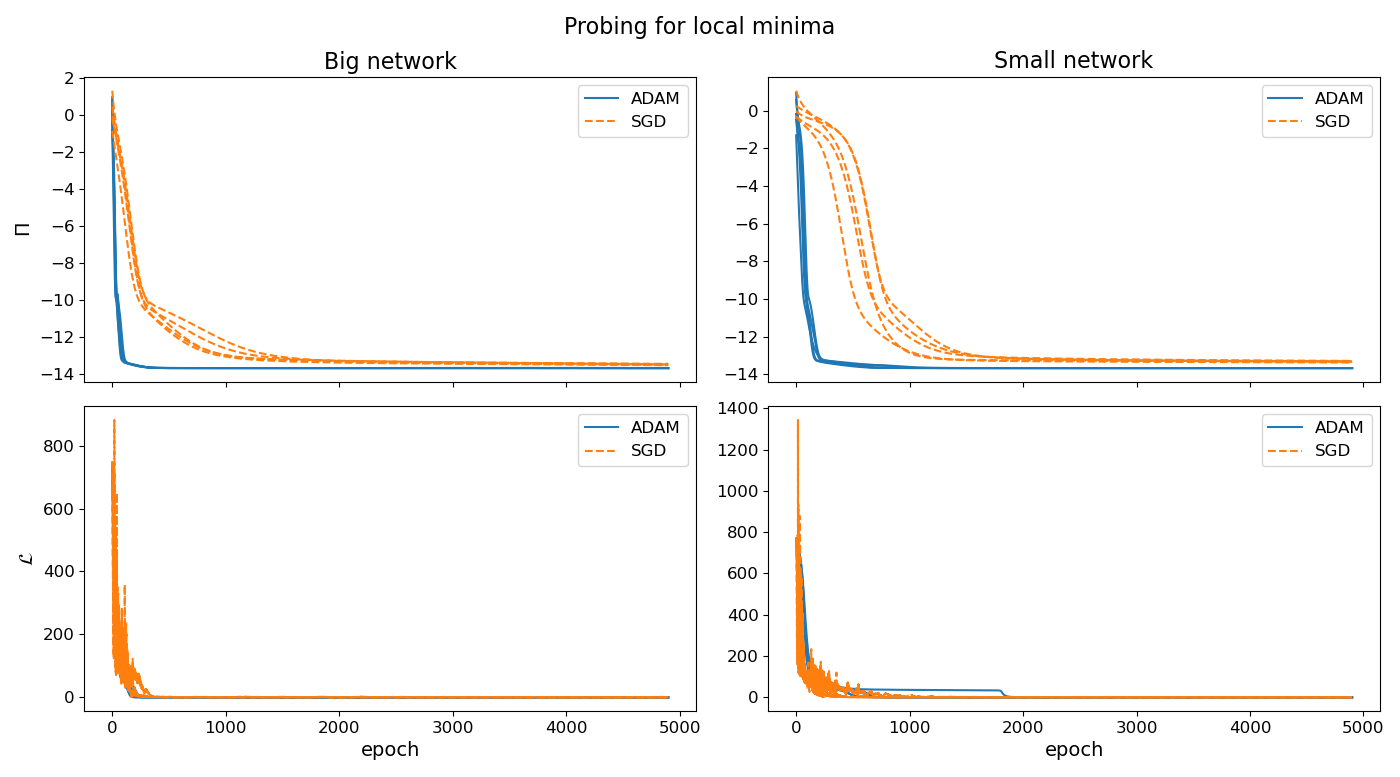}
\caption{No bad local minima are found with the DRM or PINN objective, even for small networks trained with full-batch gradient descent without momentum. We do, however, find that gradient descent obtains a slightly higher energy value at convergence for both network sizes than ADAM.}
\label{1d_probing}
\end{figure}

\paragraph{} A number of the mathematical investigations of the loss landscape show that under certain assumptions, all local minima are also global minima, meaning that the non-convexity of the loss landscape is benign \cite{kawaguchi_deep_2016, laurent_deep_2018, nguyen_loss_2017, nguyen_loss_2018, sun_global_2020}. As \cite{swirszcz_local_2017} shows, the assumptions involved in such proofs, such as linear networks and extreme overparameterization, are often not met in practice. This means that the existence of bad local minima in realistic networks must be investigated empirically. In Figure \ref{1d_probing}, we show the optimization trajectories for the DRM and PINN objectives for two different network architectures. The ``big network'' is a two hidden-layer MLP with a width of $20$, and the small network is the same architecture but with a width of $5$. The network is big in the sense that the previous examples have shown it is large enough to obtain the exact solution without getting stuck in local minima. The optimization is carried out with both ADAM optimization and gradient descent (SGD) $10$ times for each objective and network size. The step size for both optimizers is $1 \times 10^{-3}$ and the optimization is run for $5000$ epochs. Note that we abbreviate gradient descent as ``SGD'' because this is the name of the PyTorch optimizer, not because we introduce stochasticity through mini-batching the integration. We see that the network never gets stuck in a bad local minimum, even with a width of $5$. This is the case even with full-batch gradient descent without momentum, indicating that at no point does the optimizer encounter a true local minimum until the true solution is obtained. We note that the loss oscillates dramatically in the early stages of optimization with gradient descent for the PINN objective. We speculate that this is a consequence of the large values of the PINN objective and the fixed step size of vanilla gradient descent---the magnitude of $\mathcal L$ is on the order of $10^3$ at initialization but decreases rapidly, indicating that the optimizer begins in a steep region of the loss landscape. With gradient descent, the steepness directly controls the size of the parameter update, which evidently leads to steps that are too large in the early phases of optimization. On the other hand, ADAM has built-in step size control and thus avoids these oscillations. Both optimizers give rise to monotonic trajectories of the energy objective, whose gradients are small compared to the PINN objective. We note that gradient descent reliably finds a slightly larger energy value for both network sizes, though we do not consider this a bad minimum, given its proximity to the true solution. Instead, we speculate this is a consequence of flatness of the loss landscape, which causes gradient descent to halt whereas ADAM continues as a result of momentum.

\section{Two-dimensional Neohookean hyperelasticity}
\label{2d}

\paragraph{} In this section, we use the same techniques introduced above to explore the loss landscapes of DRM and PINNs for a non-linear vector-valued PDE. In particular, we work with static Neohookean hyperelasticity in two spatial dimensions. The governing equation for the Neohookean solid is

\begin{equation}\label{stress_eq}
\begin{aligned}
    \pd{P_{ij}}{X_j} + B_i = 0, \\
    \mbf u(\mbf X) = \mbf 0, \quad \mbf X \in \partial \Omega,
\end{aligned}
\end{equation}

\noindent where $\Omega$ is the reference configuration of the body, $\partial \Omega$ is the boundary, $\mbf X \in \mathbb R^2$ is the position in the reference configuration, $\mbf B$ is a body force, and $\mbf P$ is the first Piola-Kirchhoff stress tensor. Using the compressible Neohookean model of \cite{bonet_nonlinear_2008}, this stress tensor is given by

\begin{equation*}
    \mbf P = \ell_1( \mbf F - \mbf F^{-T}) + \ell_2 \log( J) \mbf F^{-T},
\end{equation*}

\noindent where $\ell_1$ and $\ell_2$ are material properties called ``Lam\'e'' constants, $F_{ij}=\delta_{ij}+\partial u_i / \partial X_j$ is the deformation gradient, and $J$ is its determinant. It can be shown that Eq. \eqref{stress_eq} corresponds to the minimum of the ``total potential energy'' functional given by 

\begin{equation}\label{tpe}
    \Pi( \mbf u(\mbf X) ) = \int_{\Omega} \qty[ \frac{\ell_1}{2} \Big( \mbf F : \mbf F - 2 - 2 \log J\Big) + \frac{\ell_2}{2}\Big( \log J\Big)^2- \mbf B \cdot \mbf u ]d\Omega.
\end{equation}

We take the reference configuration to be the unit circle centered at the origin, i.e., $\Omega=\{ (X_1,X_2): X_1^2 + X_2^2 -1 \leq 0\}$. The two displacement components are discretized with a neural network $\mathcal N: \mathbb R^2 \rightarrow \mathbb R^2$ with parameters $\bs \theta$. We again enforce the homogeneous Dirichlet boundaries with a distance-type function:

\begin{equation*}
    \mbf u(\mbf X; \bs \theta) = \phi(\mbf X) \mathcal N(\mbf X; \bs \theta),
\end{equation*}

\noindent where $\phi(\mbf X)>0$ inside the domain and $\phi=0$ along the boundary\footnote{Once again, a subsequent subsection shows that the distance function is applied at the last layer of the network, as opposed to the output. This allows the neurons in the last layer to be treated as kinematically admissible basis functions.}. In this example, we take $\phi = 1 - X_1^2 - X_2^2$. We choose to work with homogeneous Dirichlet boundaries specifically because the boundary conditions can be enforced strongly in this way, which avoids augmenting the loss function with additional terms to handle boundary conditions. We reiterate that our study of the loss landscape investigates only the differential operator acting in the bulk of the material, and does not consider the influence of modifications to the loss function for boundary condition enforcement.

\paragraph{} With the displacement discretized in terms of the neural network parameters $\bs \theta$, the loss function for DRM is simply Eq. \eqref{tpe}. The loss function for PINNs is 

\begin{equation}\label{pinn}
    \mathcal L(\bs \theta) = \frac{1}{2} \int_{\Omega} \qty(\pd{P_{ij}}{X_j} + B_i) \qty(\pd{P_{ik}}{X_k} + B_i) d\Omega.
\end{equation}

To verify our implementation of both DRM and PINNs, we use the method of manufactured solutions. In particular, with $\ell_1=1$ and $\ell_2=0.25$, we take the displacement field to be

\begin{equation}\label{neo_man}
    \mbf u(\mbf X) = \alpha(X_1^2+X_2^2)(1-X_1^2-X_2^2)\begin{bmatrix} -X_2 \\ X_1
    \end{bmatrix},
\end{equation}

\noindent which corresponds to a torsional displacement satisfying the Dirichlet boundary conditions. The twist rate is controlled by the parameter $\alpha$. This displacement field is used to compute the deformation gradient, from which the first Piola-Kirchhoff stress tensor is obtained. Then, the source term corresponding to the assumed displacement field is computed through the governing equation as 

\begin{equation*}
    B_i(\mbf X) = - \pd{P_{ij}(\mbf X)}{X_j}.
\end{equation*}

With the source term in hand, we define the DRM and PINN problems as

\begin{equation}\label{2d_objs}
    \bs \theta^{\text{DRM}}_f = \underset{\bs \theta}{\text{argmin }} \Pi(\bs \theta), \quad \bs \theta^{\text{PINN}}_f = \underset{\bs \theta}{\text{argmin }} \mathcal L(\bs \theta).
\end{equation}

\begin{figure}[hbt!]
\centering
\includegraphics[width=0.5\textwidth]{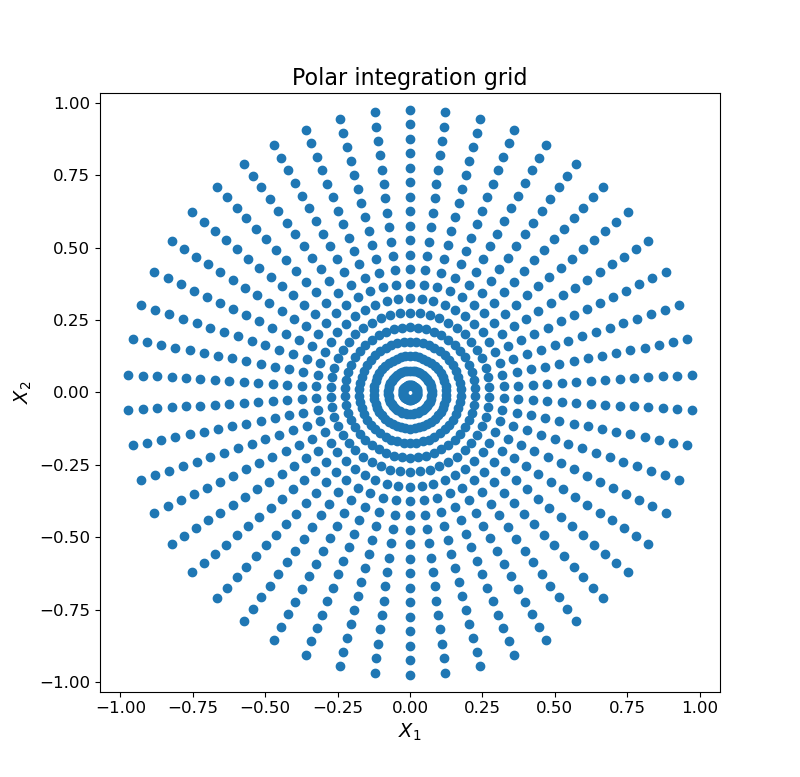}
\caption{The polar integration grid used to compute the integrals in the DRM and PINN objective functions.}
\label{polar_grid}
\end{figure}

\begin{figure}[hbt!]
\centering
\includegraphics[width=0.99\textwidth]{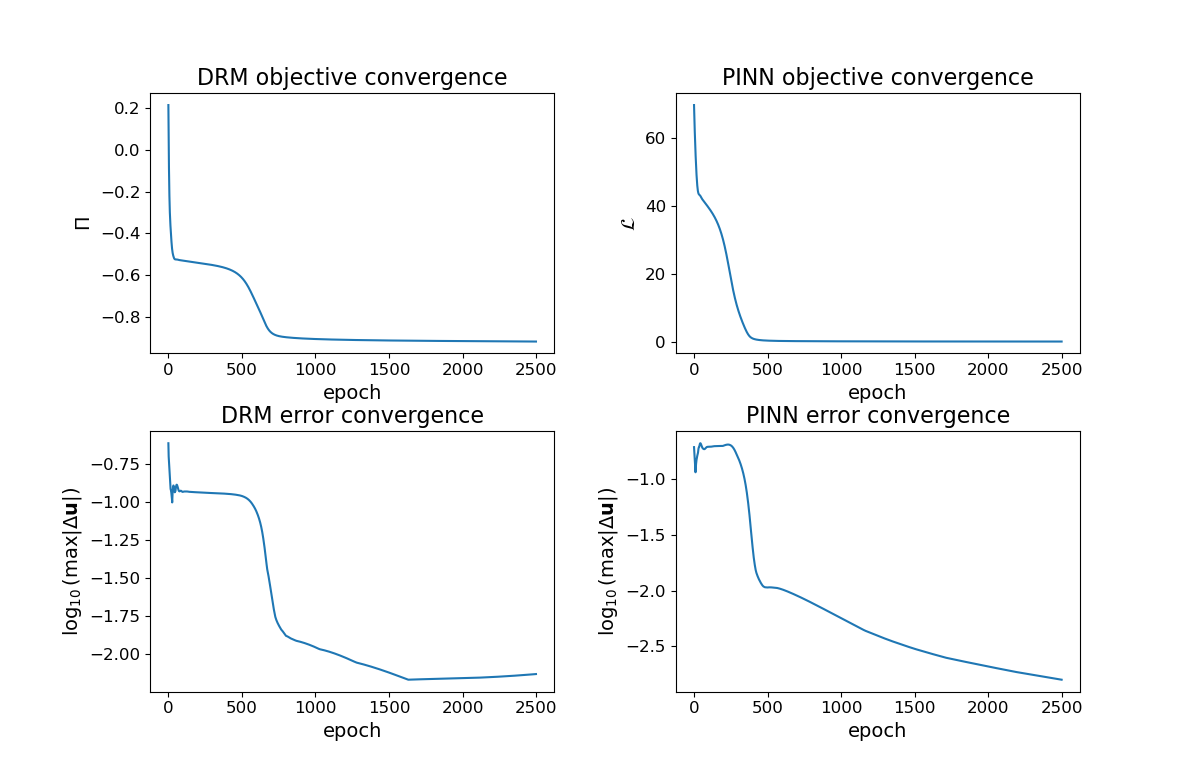}
\caption{Convergence of the DRM and PINN objectives for our given optimization settings. The maximum pointwise errors fall below $1 \times 10^{-2}$ for both methods in the allotted optimization epochs. This study is used to verify our implementation of the hyperelastic material model in the variational and strong forms.}
\label{2d_convergence}
\end{figure}

\begin{figure}[hbt!]
\centering
\includegraphics[width=0.99\textwidth]{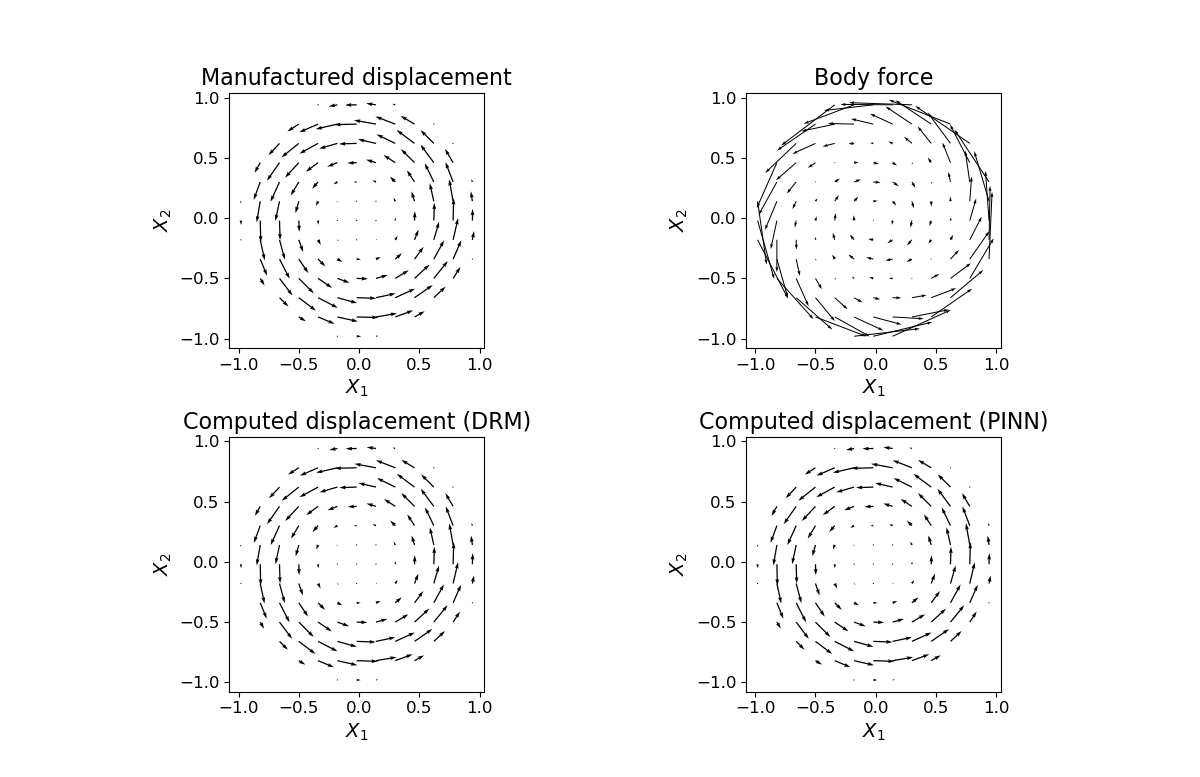}
\caption{We assume a twisting displacement field with homogeneous Dirichlet boundary conditions. The body force corresponding to this displacement field twists the hyperelastic disk clockwise near the center, and counterclockwise toward the edges. Both DRM and PINNs recover the torsional displacement field.}
\label{manufactured}
\end{figure}

 With a twist rate of $\alpha=1$, we run ADAM optimization for $2500$ epochs with a learning rate of $1 \times 10^{-3}$. The network discretizing the displacement components is a two hidden-layer MLP with width $25$ using hyperbolic tangent activation functions. To verify our implementation, we monitor the convergence of the energy and strong form loss functions over the course of optimization, as well as the maximum pointwise error with the exact solution, which we denote as $\max(|\Delta \mbf u|)$. As shown in Figure \ref{polar_grid}, the integrals in Eqs. \eqref{tpe} and \eqref{pinn} are performed with a polar integration grid with $20$ equally-spaced radial positions and $50$ equally-spaced angular positions, corresponding to $1000$ total integration points. We note from experience that a much larger Cartesian grid is required to obtain the same level of accuracy as this polar grid. See Figure \ref{2d_convergence} for the convergence of the two objectives and maximum errors. For both objectives, the maximum pointwise error is less than $1 \times 10^{-2}$ at convergence. Figure \ref{manufactured} shows the manufactured displacement field, the corresponding body force, and the displacement fields computed from DRM and PINNs. 

\subsection{Monotonic linear interpolation}

\begin{figure}[hbt!]
\centering
\includegraphics[width=0.99\textwidth]{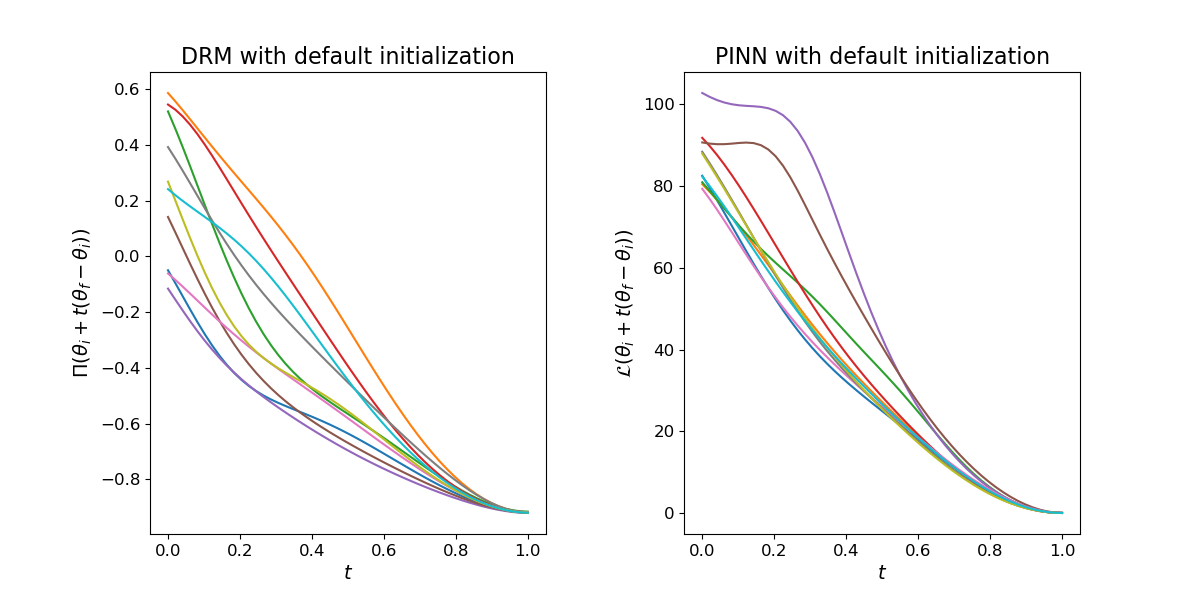}
\caption{Even with the nonlinear hyperelastic constitutive model, straight-line paths from the initial to the final parameters tend to decrease monotonically the loss function for both DRM and PINNs. There are, however, two paths with the PINN objective that do not follow this trend, one of which has a flat region and the other shows a small barrier along the straight-line path.}
\label{2d_MLI}
\end{figure}

\paragraph{} We repeat the same MLI analysis carried out in the previous section for the two hyperelastic loss functions. However, we cannot arbitrarily scale the initialization, as initializing the parameters to be too large gives rise to displacement fields which ``invert'' the material. This means that the determinant of the deformation gradient is negative, and thus its logarithm, which is required in the constitutive model, is not defined. In theory, this is possible with the default initialization, but we only observe inversion when the initial parameters are artificially scaled up. Thus, we randomly initialize the parameters $10$ times for both objectives, compute solutions with Eq. \eqref{2d_objs}, and plot the loss along a straight-line path joining the initial and final parameters. The results of this are shown in Figure \ref{2d_MLI}. In the majority of the $10$ trials for the two objectives, the loss monotonically decreases on the path connecting the initial and final parameters. This is the case despite the increased complexity of the hyperelastic loss functions. However, with the PINN objective, two paths show barriers in the linear interpolation between the initial and final states.

\subsection{Exploring the solution manifold}

\begin{figure}[hbt!]
\centering
\includegraphics[width=0.99\textwidth]{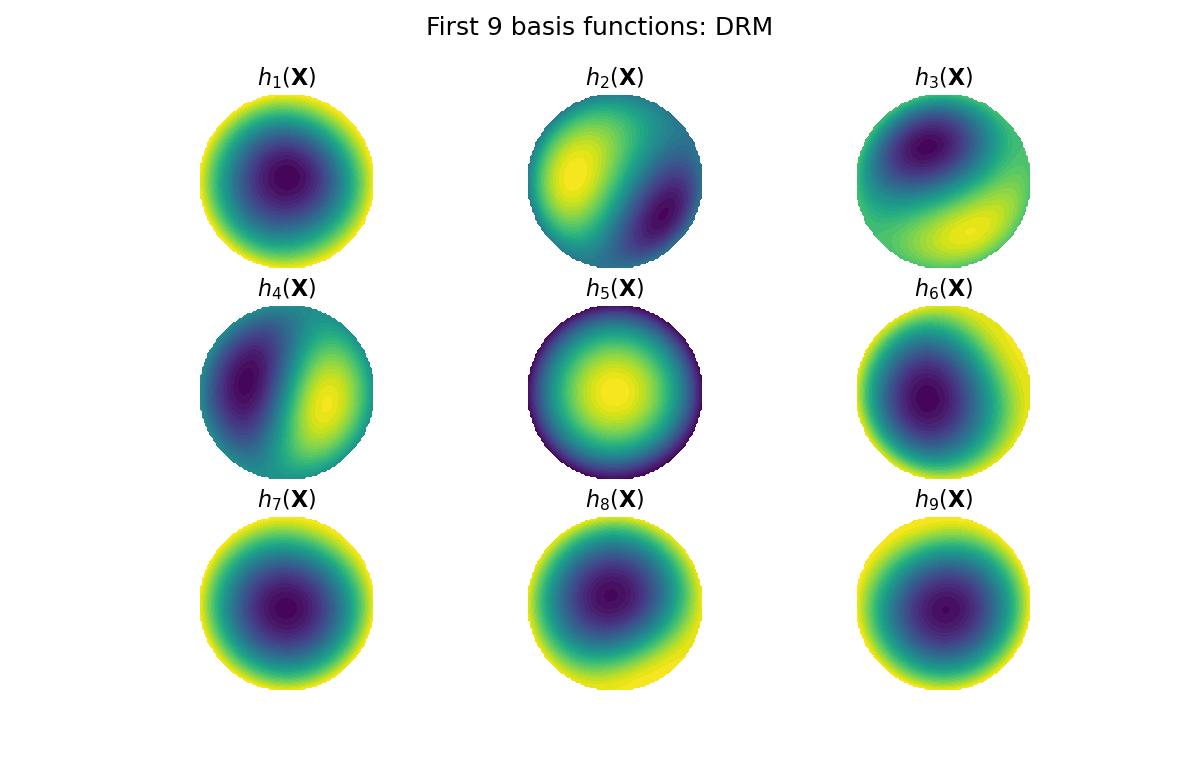}
\caption{Plotting $9$ of the $25$ basis functions learned by training the network with the DRM objective. The redundancy in the basis suggests that only a subset of these functions is required to fit the solution.}
\label{2d_drm_basis}
\end{figure}

\begin{figure}[hbt!]
\centering
\includegraphics[width=0.99\textwidth]{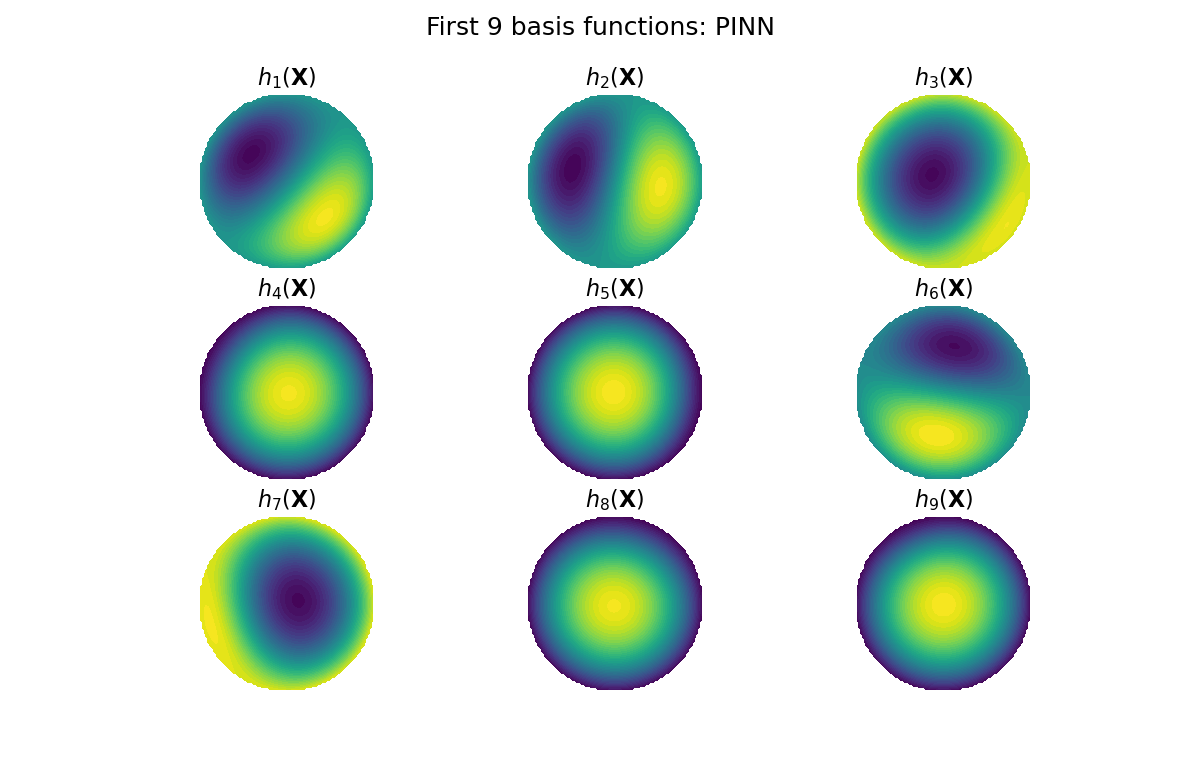}
\caption{Plotting $9$ of the $25$ basis functions learned by training the network with the PINN objective. This basis has the same qualitative features as the basis obtained from DRM.}
\label{2d_pinn_basis}
\end{figure}

\paragraph{} In the case of the one-dimensional elliptic problem, we found that taking steps in the directions of Hessian eigenvectors with zero associated eigenvalues was one way to explore the manifold of local minima. Later, we showed that when the learned basis functions are linearly dependent, there are guaranteed to be linear spaces of equivalent solutions. In other words, linear dependence of the basis functions gives rise to planes of equal loss whose dimension is controlled by the rank of the matrix of basis functions. In this example, we explore the solution manifold purely from the perspective of dependence of the basis functions. Given the manufactured solution of Eq. \eqref{neo_man}, the network needs only to learn two relatively simple basis functions: $(X_1^2+X_2^2)(1-X_1^2-X_2^2)X_1$ and $(X_1^2+X_2^2)(1-X_1^2-X_2^2)X_2$. Of course, this is not the only way to accurately represent the solution field, but we point this out in order to understand how many of the basis functions could be redundant. We note that the neural network discretization of the displacement field has the following form:

\begin{equation*}
    \mbf u(\mbf X; \bs \theta) = \sum_{i=1}^N \bs \theta_i^O (1-X_1^2-X_2^2)\tilde h_i( \mbf X; \bs \theta^I),
\end{equation*}

\noindent where $\bs \theta_i^O$ are the outer parameters---which are now vectors---mapping basis functions $h_i(\mbf X) = (1-X_1^2-X_2^2)\tilde h_i( \mbf X)$ to the displacement components. Using the two hidden-layer network of width $25$ and ADAM optimization with a learning rate of $1 \times 10^{-3}$ for $3000$ epochs, we train two networks representing the DRM and PINN solutions to convergence. In Figures \ref{2d_drm_basis} and \ref{2d_pinn_basis}, $9$ out of the $25$ learned basis functions are shown. We remark that in both cases, functional forms resembling both of the required basis functions are seen. To be clear, these are the basis functions with opposite sign on either side of a line equally dividing the circular domain, which are equivalent up to a rigid rotation. Visually, the basis functions show a large degree of redundancy, as many have a ``bubble'' shape that contributes negligibly to the torsional displacement field.

\paragraph{} As before, we form the matrix of basis functions around the solution parameters $\bs \theta^f$. This is given by 

\begin{equation*}
    \mbf H(\mbf X) = \begin{bmatrix}
         h_1(\mbf X; \bs \theta^{f,I}) &  h_2(\mbf X; \bs \theta^{f,I}) & \dots &  h_N(\mbf X; \bs \theta^{f,I})
    \end{bmatrix}.
\end{equation*}

\begin{figure}[hbt!]
\centering
\includegraphics[width=0.99\textwidth]{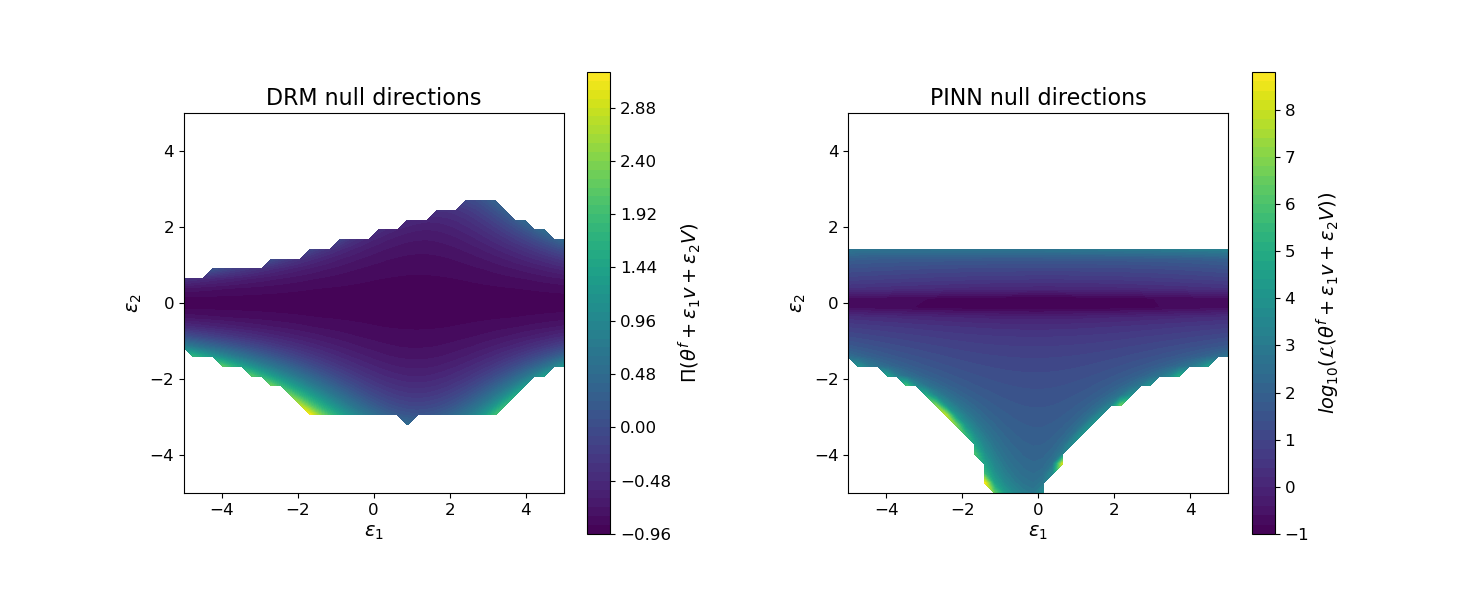}
\caption{When the network learns linearly dependent basis functions, the loss landscape has trenches of equivalent loss which extend to infinity. We remark that learning linearly independent basis functions is a challenging problem, and should not be expected in general \cite{rowan_solving_2025}.}
\label{2d_null}
\end{figure}

We then compute the Gram matrix $G_{ij} = \int_{\Omega} h_i(\mbf X) h_j(\mbf X) d\Omega$ and determine its rank. In this case, the rank is determined by counting the number of eigenvalues greater than $10^{-6}$ times the maximum. For DRM, the rank of the Gram matrix is $14$ and for PINN, it is $15$. Thus, the eigenvector of the Gram matrix corresponding to the minimum eigenvalue, which we denote as $\bs \Delta$, can be used to find an infinite flat direction in parameter space. In particular, this direction is given by $\mbf v = [\bs \Delta , \mbf 0]$, which corresponds to keeping a fixed basis and moving the outer parameters in the nullspace of the matrix of basis functions. To visualize these null directions, we plot the DRM and PINN loss functions in a plane spanned by $\mbf v$ and a randomly chosen direction $\mbf V$. Figure \ref{2d_null} shows the same ``trench'' in the loss landscape we observed in the previous example. In the case of hyperelasticity, parameter settings corresponding to unphysical displacements cause the objective to return $\texttt{nan}$. Thus, we show only the patches of parameters which yield a physical displacement field and thus a finite objective value.

\subsection{Mode connectivity}

\begin{figure}[hbt!]
\centering
\includegraphics[width=0.99\textwidth]{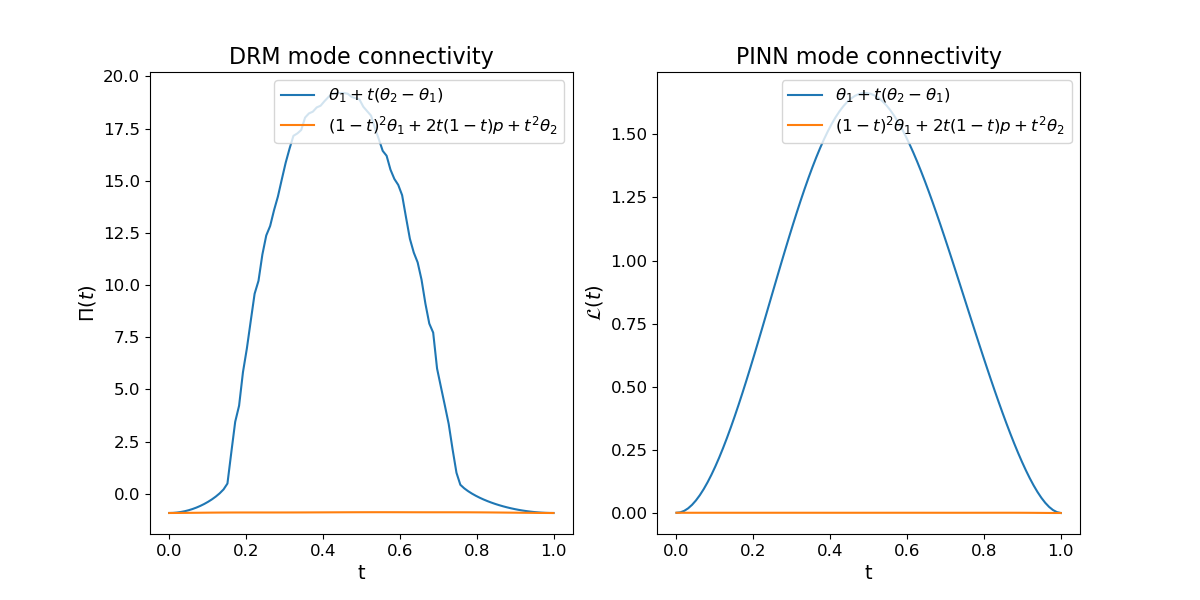}
\caption{When two distinct solutions are obtained using close initializations, a barrier separates these solutions with a linear connection, but a curved path can be found to connect them without increasing the loss.}
\label{2d_mode_connectivity}
\end{figure}

\paragraph{} We study the connectivity of different solutions of the DRM and PINN hyperelastic loss functions. Two solutions for each loss function are obtained by solving the corresponding optimization problem  with two random initializations. The two solutions, denoted $\bs \theta_1$ and $\bs \theta_2$, are first connected with a straight-line path. We find that the linear connection generally includes displacement fields which invert the material. We thus clamp the determinant of the deformation gradient at $10^{-6}$ in order to avoid logarithms of zero or negative numbers. This ensures that the loss function over the linear connection is finite, which is useful for the sake of visualization. We then solve the optimization problem of Eq. \eqref{mode_connectivity} for the control point $\mbf p$ of the quadratic Bezier curve. See Figure \ref{2d_mode_connectivity} for the results. With this single control point, a path can be found over which the loss does not increase. We note, however, that finding this optimal control point with the hyperelastic DRM and PINN objectives is a challenging optimization problem, which often fails to converge. Our results here were obtained by taking one random initialization, and then perturbing it with random normal noise to obtain a second initialization. The standard deviation of the i.i.d. random noise added to the first initialization was set at $0.1$. Larger perturbations to the first initialization lead to optimization problems for the control point that fail to converge. Further work is required to understand whether this is a failure of the optimization process---which is challenging due to the highly non-linear physics and the clamped logarithms---or a genuine failure of the mode connectivity property to hold.

\subsection{Random directions}

\begin{figure}[hbt!]
\centering
\includegraphics[width=0.99\textwidth]{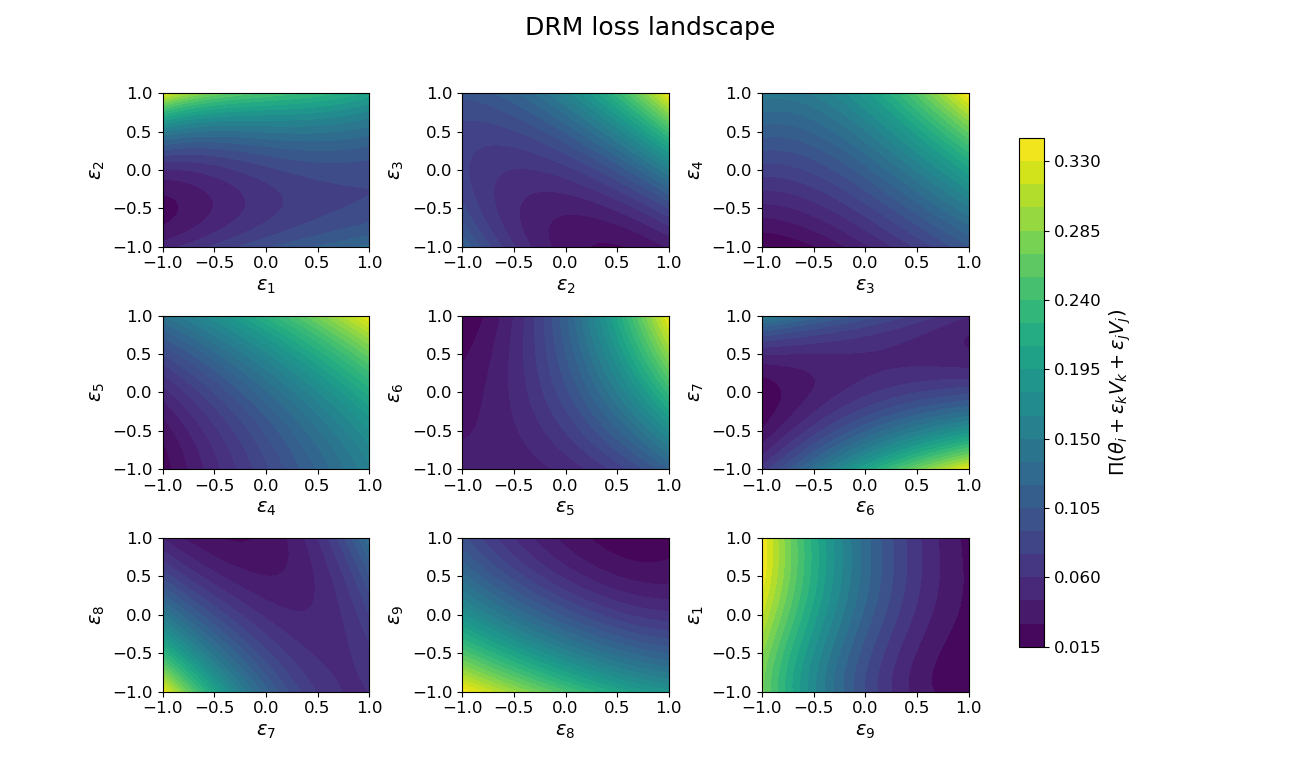}
\caption{Visualizing the DRM loss landscape around randomly initialized parameters in planes spanned by $9$ combinations of random vectors. The loss is smooth, but appears to contain more saddle points than the comparatively simple one-dimensional example problem.}
\label{2d_drm_random}
\end{figure}

\begin{figure}[hbt!]
\centering
\includegraphics[width=0.99\textwidth]{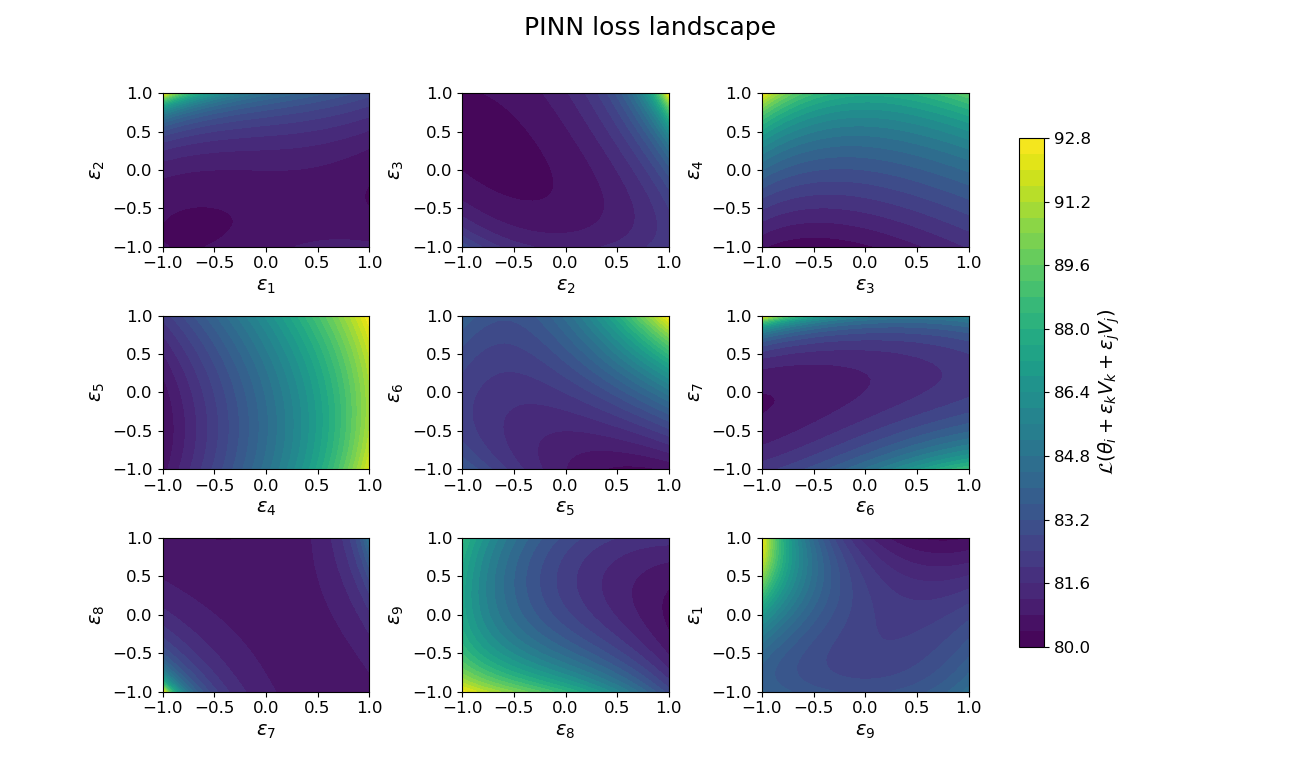}
\caption{Visualizing the PINN loss landscape around randomly initialized parameters in planes spanned by $9$ combinations of random vectors. By comparing the DRM and PINN loss landscapes, one can see that, to a large extent, the gradient directions agree. This suggests that the two formulations of the loss behave similarly from the perspective of optimization.}
\label{2d_pinn_random}
\end{figure}

\begin{figure}[hbt!]
\centering
\includegraphics[width=0.99\textwidth]{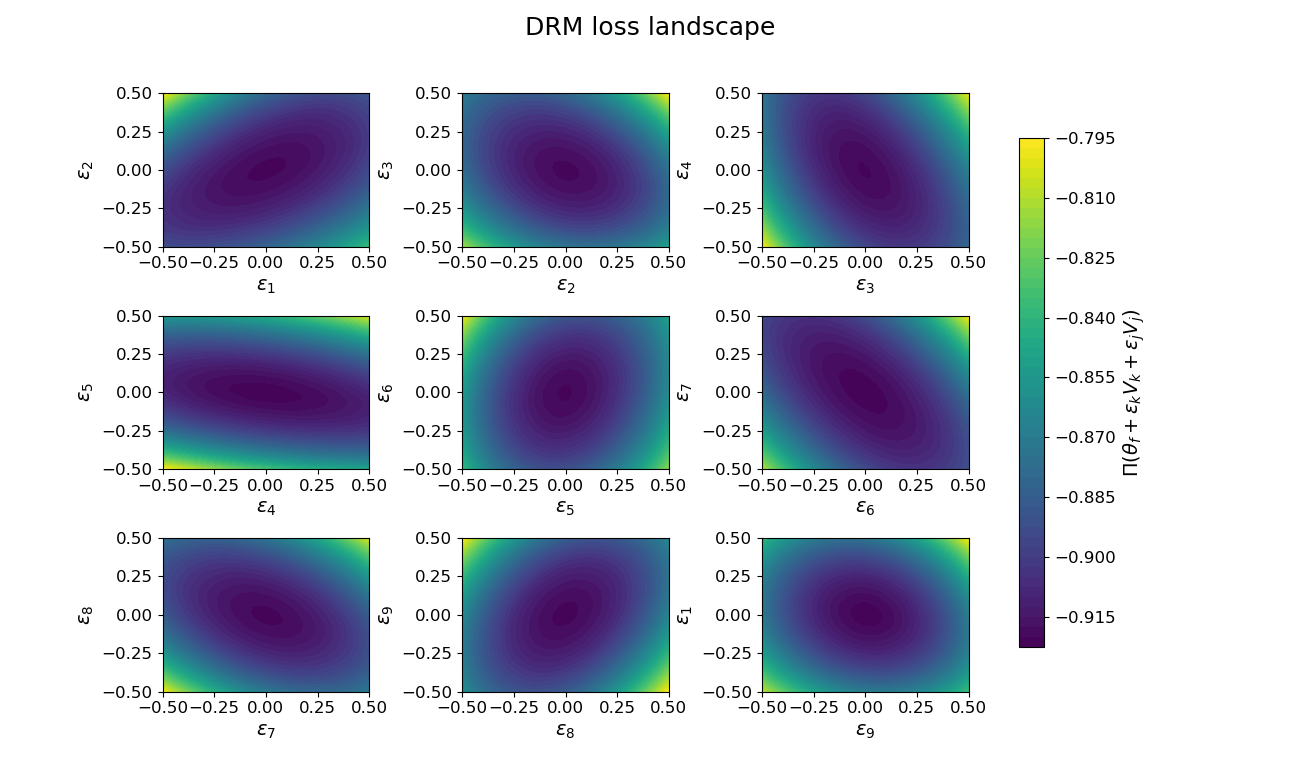}
\caption{Around the solution, the DRM loss landscape is smooth, convex, and well-conditioned when viewed in random directions.}
\label{2d_drm_random_sol}
\end{figure}

\begin{figure}[hbt!]
\centering
\includegraphics[width=0.99\textwidth]{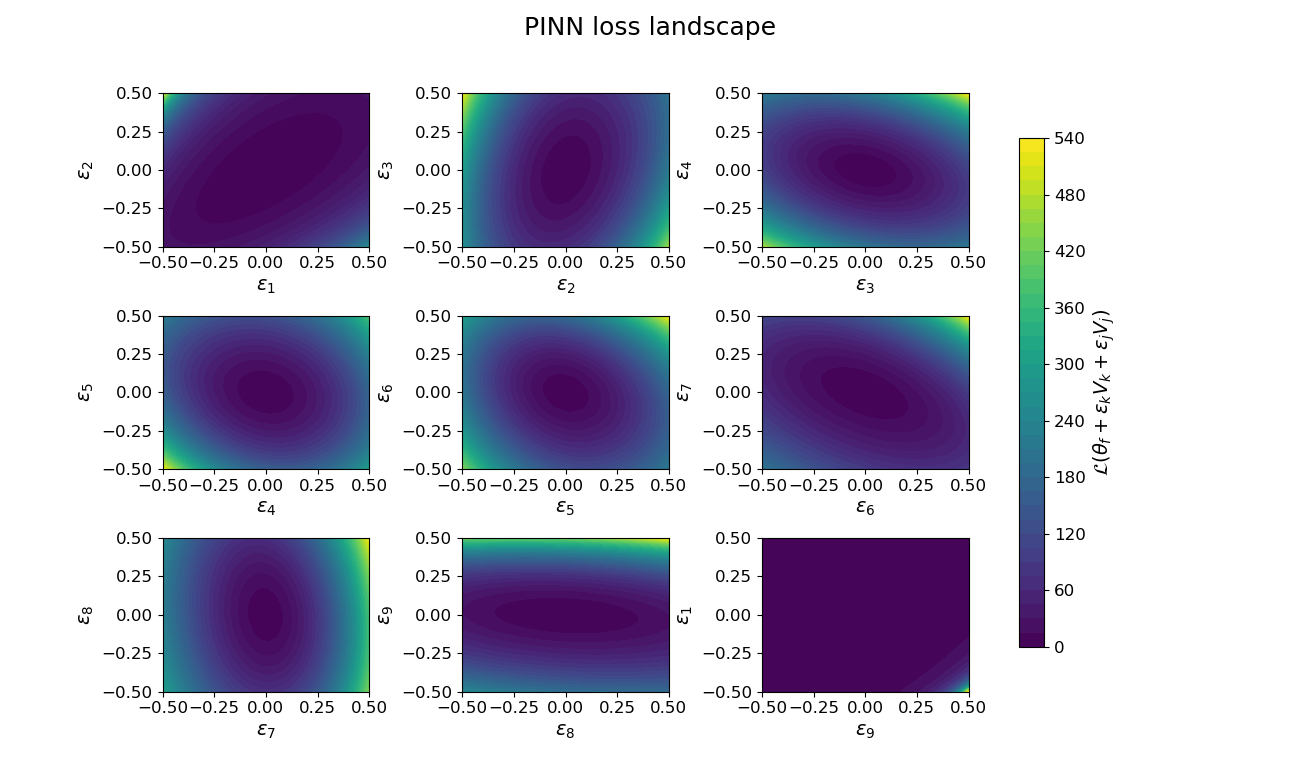}
\caption{Around the solution, the PINN loss landscape is smooth, convex, and well-conditioned when viewed in random directions.}
\label{2d_pinn_random_sol}
\end{figure}

\paragraph{} Here, we view cross-sections of the hyperelastic loss functions in planes spanned by random vectors. As before, we generate contour plots with $9$ combinations of random directions for both objectives. Figure \ref{2d_drm_random} and \ref{2d_pinn_random} show the loss surface around a randomly initialized point $\bs \theta_i$. We note that this random point is the same for both DRM and PINNs, so that all differences in the observed contours are due to the differences in the formulations of the loss function. Compared to the one-dimensional elliptic problem, the hyperelastic loss shows more signs of non-convexity through saddle points. However, as seen in Figures \ref{2d_drm_random_sol} and \ref{2d_pinn_random_sol}, the loss function around the solution is smooth and convex in the random planes. Both of our example problems illustrate that around the solution, the neural network loss landscape is well-behaved from an optimization perspective. This is the case in the random directions we probe, at least. We believe the observed smoothness and convexity of the loss to run counter to the intuitions of many ML practitioners. As an example, in \cite{kiyani_optimizing_2025}, the authors say that many optimizers struggle in the ``highly non-linear and non-convex loss landscapes, leading to challenges such as slow convergence, local minima entrapment, and non-degenerate saddle points.''

\paragraph{} We also note that when going to a three hidden-layer network of the same width, the same qualitative features in the loss landscape are observed, except that smaller parameter changes have more drastic effects on the displacement and thus invert the material more readily. From our experiments, the loss does not take on the chaotic character noted in \cite{li_visualizing_2018}, but the networks we use are also not as wide or deep as those used in the data-driven classification problems studied in that work.

\subsection{Hessian eigenvalues}

\begin{figure}[hbt!]
\centering
\includegraphics[width=0.99\textwidth]{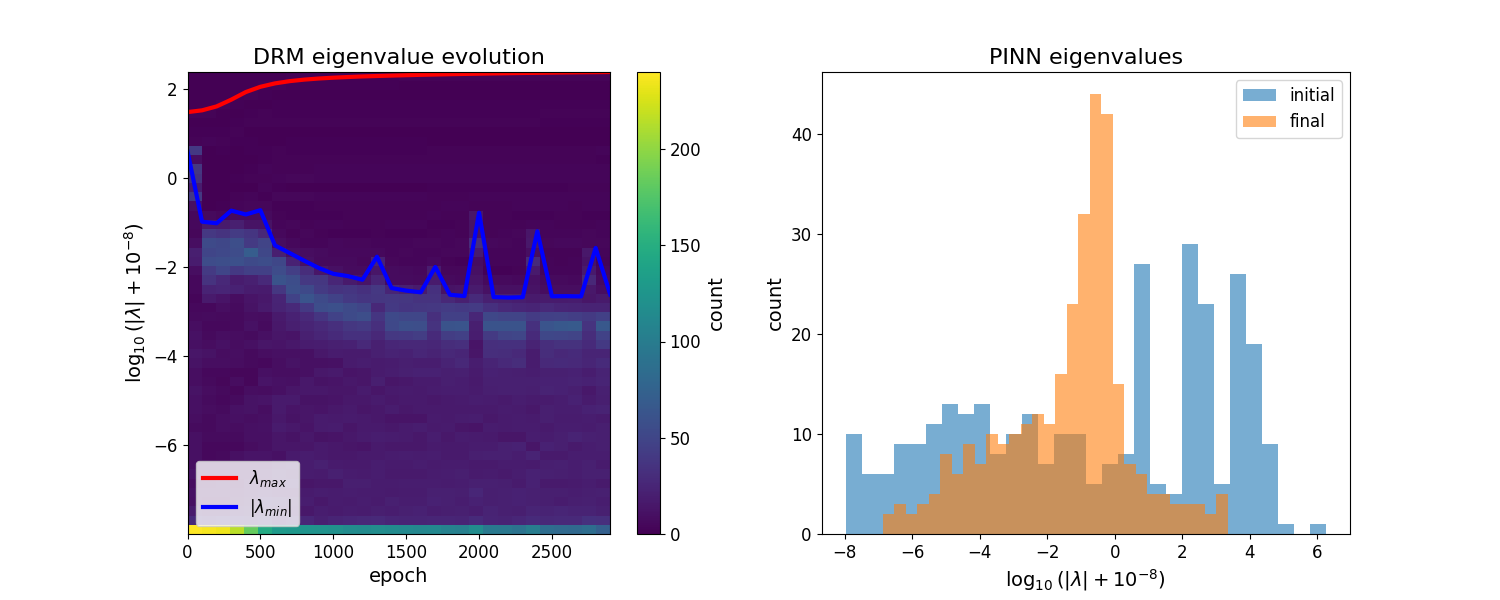}
\caption{From the perspective of the Hessian computed along the training trajectory, the two-dimensional hyperelastic and the one-dimensional linear DRM loss functions behave similarly. The optimizer leaves the non-convex region around initialization, entering into a flat basin with a small number of increasingly steep directions.}
\label{2d_eigenvalues}
\end{figure}

\paragraph{} We now perform eigenanalysis on the Hessian matrices from the DRM and PINN problems. With the one-dimensional elliptic problem, we plotted the evolution of the Hessian eigenvalues over the course of optimization. This provided insight into how the geometry of the loss landscape in the vicinity of the parameters evolved under the optimization dynamics. We find that conducting a similar analysis on the PINN objective is infeasible. To compute the Hessian, we need a matrix of second derivatives of the loss with respect to the parameters. This requires differentiating through the square of the strong form loss, which itself involves second spatial derivatives of the displacement components and is highly non-linear. On two separate machines, PyTorch crashes when attempting to use automatic differentiation to compute the Hessian of the PINN loss for a two hidden-layer network of width $25$. However, the DRM formulation of the problem presents no such challenge, likely owing to the lower order of differentiation. Thus, we plot the eigenvalue evolution of the DRM loss landscape, and show only the initial and final eigenvalues of the Hessian of the PINN loss, but using a network of width $15$. The results are given in Figure \ref{2d_eigenvalues}. Qualitatively, the evolution of the eigenvalues in this problem is equivalent to the one-dimensional example. The maximum eigenvalue increases and stabilizes, the negative curvature disappears, the majority of eigenvalues are concentrated around zero, and a band of moderate-sized eigenvalues decays away. The initial and final eigenvalues of the PINN objective suggest a similar trend: the eigenvalue spectrum is more concentrated around zero by the end of training.

\subsection{Goldilocks zone}

\begin{figure}[hbt!]
\centering
\includegraphics[width=0.99\textwidth]{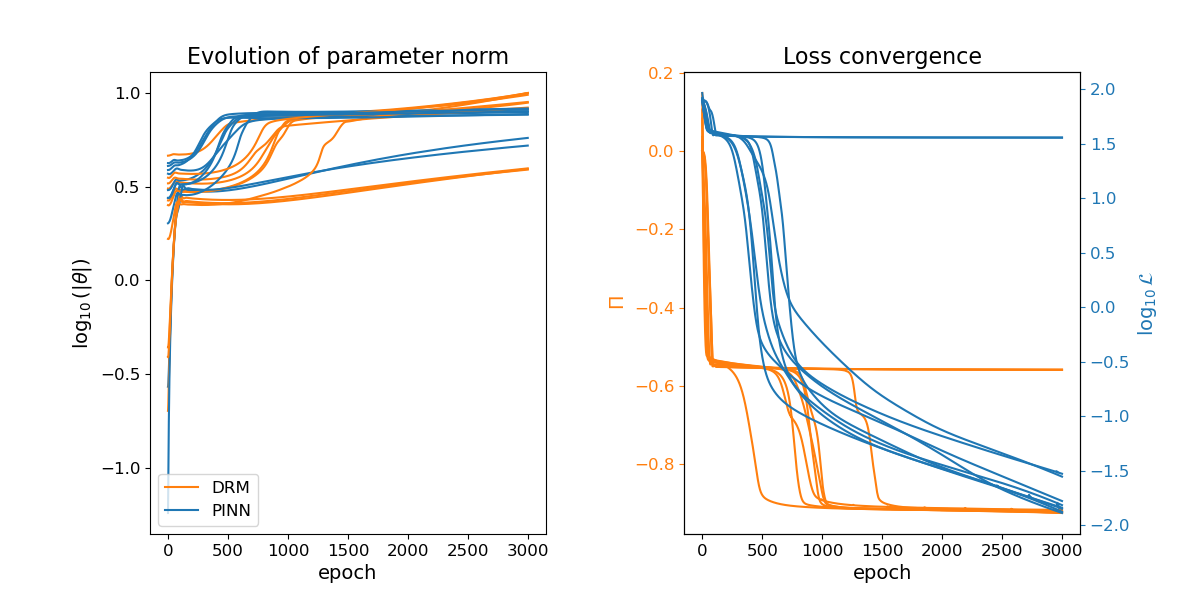}
\caption{Similar to the one-dimensional example problem, small parameter magnitudes increase to a roughly constant value over the course of training, independent of their initialization. In $4$ of the $20$ trials, the parameters do not make it to this region, and the network does not obtain the lower-valued minimum of the other trials.}
\label{2d_goldilocks}
\end{figure}

\paragraph{} In this example, we investigate the relationship between the norm of the parameters and the success of training. With hyperelasticity, we avoid training on the surface of hyperspheres of different radii because large parameters correspond to displacements that invert the material. Instead, we randomly initialize the parameters, and then randomly choose the magnitude by multiplying the initial parameters by a standard uniform random variable. This ensures that the parameter magnitude varies, but avoids the unphysical displacement fields from large parameter settings. We do this for $10$ samples with both the DRM and PINN loss. See Figure \ref{2d_goldilocks} for the results. Regardless of the initial state, the parameter magnitude approaches the Goldilocks region we observed in the previous example, which corresponds to $|\bs \theta|\approx 10$. This is the case for all runs except for $2$ from DRM and $2$ from the PINN loss. We note that the four networks whose parameters do not reach the Goldilocks zone are the same which do not achieve low loss values. Furthermore, the point around which the energy stagnates appears to be the same local minimum observed in a subsequent subsection.

\subsection{Intrinsic dimensionality}

\begin{figure}[hbt!]
\centering
\includegraphics[width=0.99\textwidth]{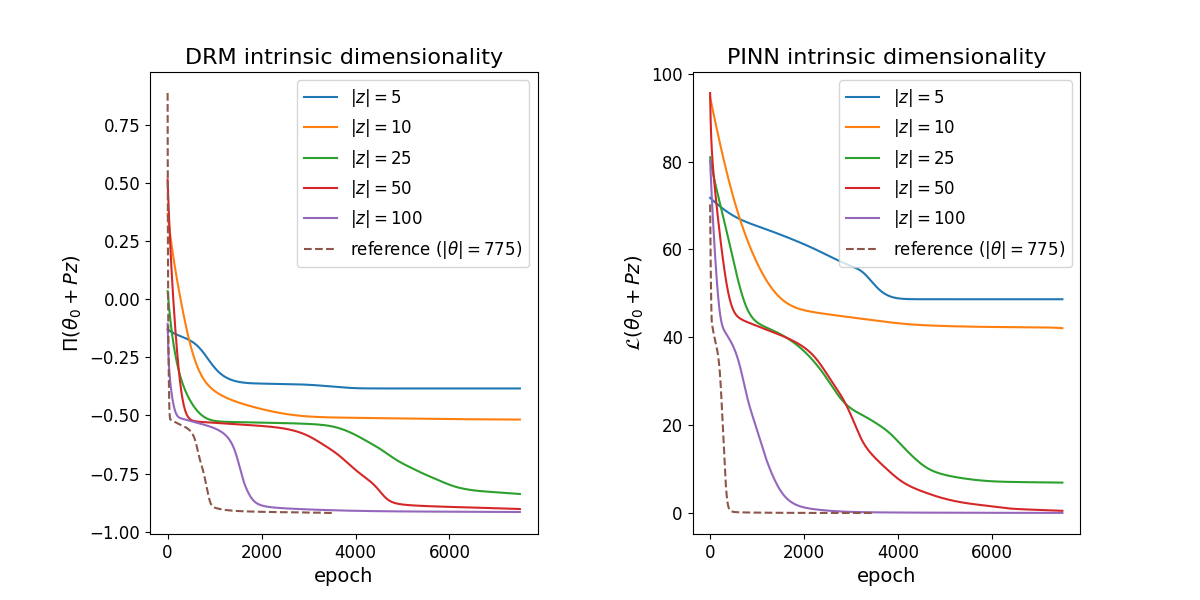}
\caption{As expected, increasing the dimension of the random subspace leads to monotonic improvements in performance, as measured by the converged loss value. With only $50$ parameters, both networks achieve similar loss values to the reference solution, which is obtained with a network of $775$ parameters.}
\label{2d_intrinsic_dimensionality}
\end{figure}

\paragraph{} To investigate the existence of low-dimensional structure in the network, we train the DRM and PINN loss functions in a random affine subspace of the parameters. As before, the training problems become 

\begin{equation*}
    \underset{\mbf z^{\text{DRM}}}{\text{argmin }} \Pi( \bs \theta^{\text{DRM}}_0 + \mbf P \mbf z^{\text{DRM}}), \quad \underset{\mbf z^{\text{PINN}}}{\text{argmin }} \mathcal L( \bs \theta^{\text{PINN}}_0 + \mbf P \mbf z^{\text{PINN}}),
\end{equation*}

\noindent where $\mbf P$ is a random projection matrix. Like the previous example, we take the parameters defining the offset of the affine subspaces to be equal, and to come from PyTorch's default initialization. We sweep over $5$ subspace dimensions and train in each subspace until convergence. We use the same two hidden-layer network of width $25$ with ADAM optimization with a learning rate of $1 \times 10^{-3}$. The projection matrix is the same between DRM and PINN at each subspace dimension but is re-initialized when the subspace dimension changes. See Figure \ref{2d_intrinsic_dimensionality} for the results. As the dimension of the subspace increases, the converged value of the energy and strong form loss decreases. While the network has $775$ parameters, we obtain approximately equivalent performance, as measured by the final loss value, for a random subspace of dimension $50$. Our two sets of results on intrinsic dimensionality suggest that the networks are overparameterized, even in spite of their modest size. It is common in the SciML literature to use networks with thousands of parameters, as opposed to the hundreds that we use here. As discussed in \cite{rowan_possibility_2026}, very small networks often suffice for problems of practical interest.

\subsection{Optimization trajectory}

\begin{figure}[hbt!]
\centering
\includegraphics[width=0.99\textwidth]{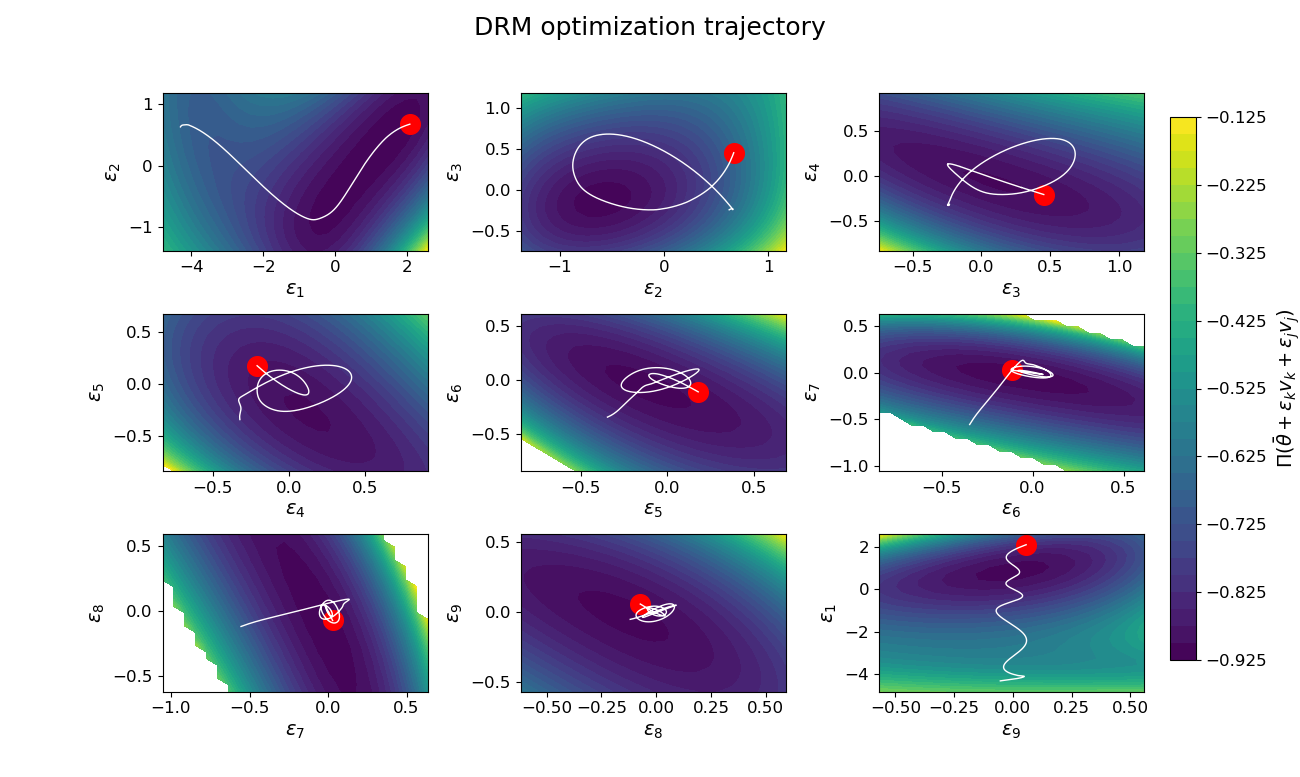}
\caption{Projecting the DRM optimization dynamics into planes spanned by combinations of principal components. The path through parameter space is well-approximated in only two dimensions. In this first principal plane, the loss landscape is remarkably simple.}
\label{2d_drm_traj}
\end{figure}

\begin{figure}[hbt!]
\centering
\includegraphics[width=0.99\textwidth]{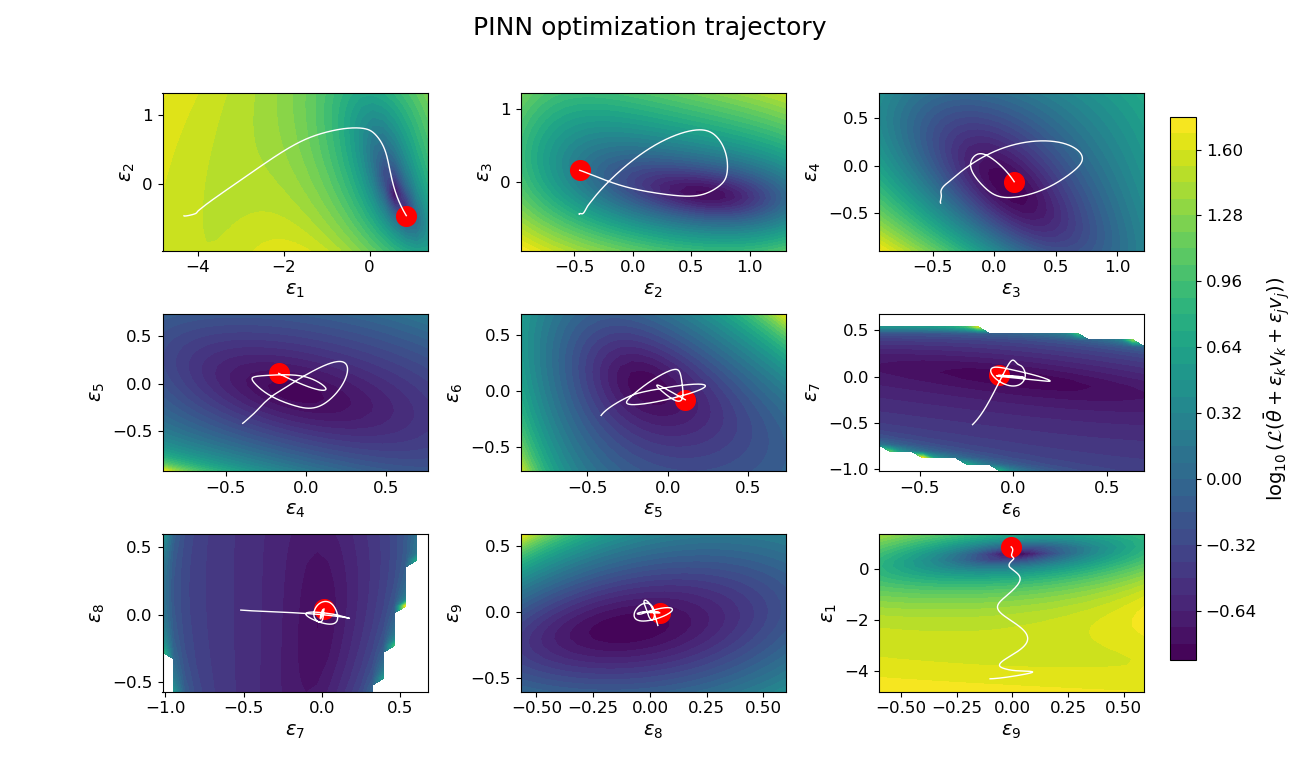}
\caption{Projecting the PINN optimization dynamics into planes spanned by combinations of principal components. Like DRM, the loss landscape is simple in the first principal plane. We find that the logarithm of the loss is more instructive here, given the large values the PINN loss takes on. Interestingly, the first principal plane provides some insight as to why the MLI property may fail to hold for the PINN objective, as a small barrier is encountered along the straight line between initial and final projected states.}
\label{2d_pinn_traj}
\end{figure}

\paragraph{} We train two networks on the DRM and PINN problems, store the parameters over the course of the training process, and perform principal component analysis on this data. By looking at the fraction of the variance explained by each principal component, we gain insight into whether training occurs in a low-dimensional subspace. For DRM, the first two principal components explain $91.8\%$ and $4.7\%$ of the variance in the trajectory data respectively. For PINNs, the first two principal components explain $90.0\%$ and $6.2\%$ of the variance. Figures \ref{2d_drm_traj} and \ref{2d_pinn_traj} show the optimization dynamics projected onto the principal planes. Recall that $\bar{\bs \theta}$ is the average of the trajectory data for each method, which defines the center of the contour plots. For DRM and PINNs, the plane spanned by the first two principal components explains more than $95\%$ of the optimization dynamics. The loss is smooth, convex, and well-conditioned in this plane. This suggests that, despite the nonlinear physical model defining the optimization problem, the loss landscape is rather simple from the perspective of the optimizer.

\subsection{No bad local minima}

\begin{figure}[hbt!]
\centering
\includegraphics[width=0.99\textwidth]{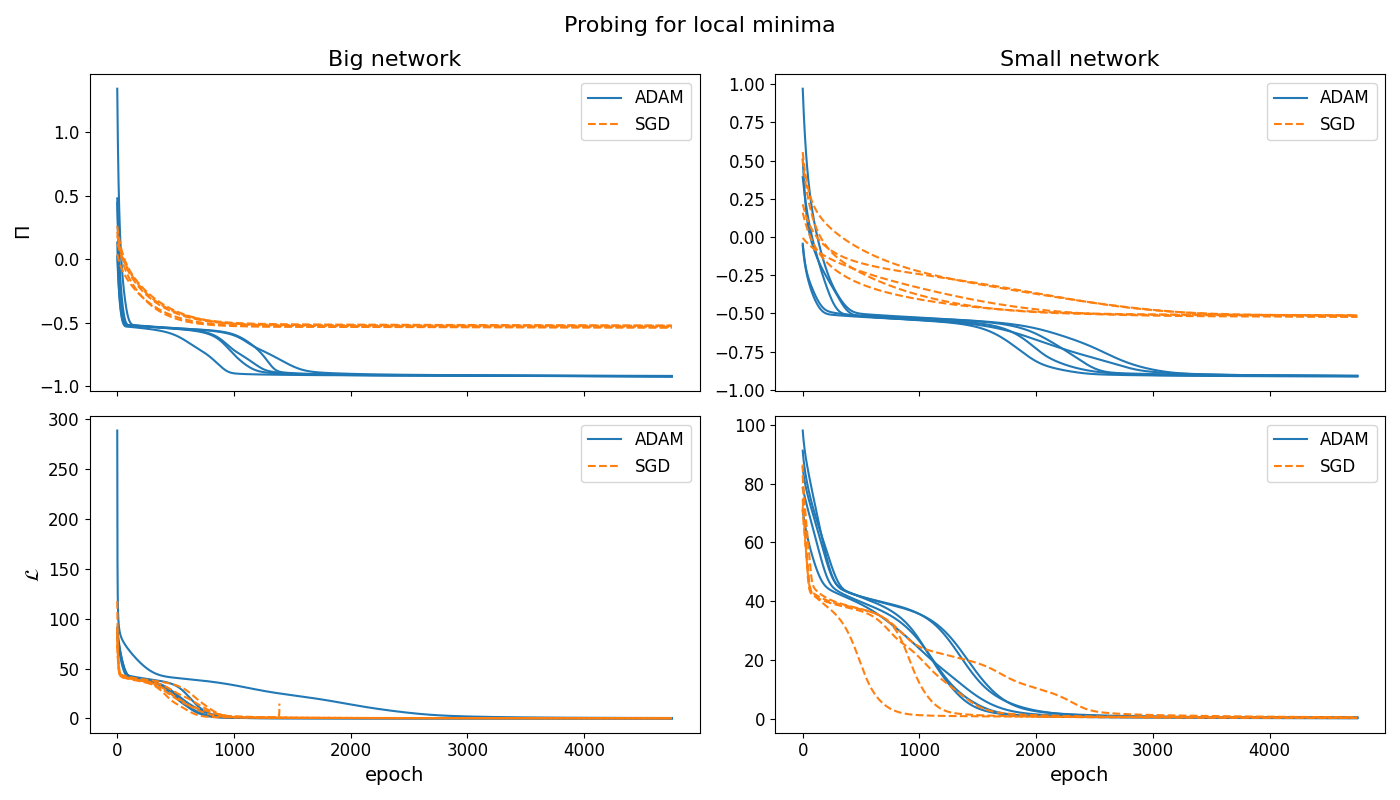}
\caption{Probing the loss landscape for bad local minima by repeatedly training networks of different sizes with different optimizers. We use gradient descent without momentum in addition to ADAM, as we expect this optimizer to be more prone to get stuck in local minima. We find that gradient descent stagnates before obtaining the true solution with the energy objective.}
\label{2d_probing}
\end{figure}

\paragraph{} As with the one-dimensional example problem, we probe the loss landscape to determine if the optimizer encounters bad local minima. We do this with both ADAM and SGD, and two different network sizes. The ``big'' network is the one we have been using up until now---a two hidden-layer MLP of width $25$. The small network is the same two hidden-layer architecture, but a width of $8$ neurons. The learning rate is decreased to $5 \times 10^{-4}$ in order to avoid inverting the material in repeated runs, which occurs more often with gradient descent. We initialize $5$ networks for each objective, each optimizer, and each network size. We then run optimization for $5000$ epochs, which is observed to be sufficient to obtain convergence in all cases. Figure \ref{2d_probing} shows the results of this study. Unlike the previous example, gradient descent gets stuck in local minima for both big and small networks when minimizing the energy objective. This is not the case, however, for the strong form objective. At this point, it is not clear what differences in the loss landscapes from the two objectives explain this. From a practitioner's perspective, this example illustrates the usefulness of ADAM over SGD, as its performance is more robust to local minima and flat regions in the loss landscape.

\section{Conclusion}
\label{conclusion}

\paragraph{} In this work, we reviewed techniques for loss landscape visualization from the literature on ML for data-driven classification problems, and then applied these techniques to two examples from physics-informed machine learning. We also introduced some original contributions, such as the ``Hessian walk'' for exploring the solution manifold, and the ``acceleration'' of the parameters as a measure of the curvature of the optimization path. Furthermore, the fact that the parameters of a physics-informed network can be broken down into ``inner'' and ``outer'' components---the former building basis functions and the latter scaling them---allowed us to interpret one source of singularity of the Hessian matrix. The primary purpose of this work has been to bring existing visualization techniques into the SciML community. A secondary purpose has been to compare the loss landscapes arising from the variational and strong form loss functions. Generally, our conclusions have been that physics-informed loss landscapes have many of the same properties as the loss landscapes of other ML problems: monotonic linear interpolation, mode connectivity, singular Hessian matrices, preferred radii in parameter space, low intrinsic dimensionality, and low-dimensional training dynamics. We have also found, at least in random planes, that the physics-informed loss landscape is smooth, convex, and well-conditioned in the vicinity of the solution. This challenges the prevailing intuition that loss landscapes are noisy and ill-conditioned. That being said, we remark that boundary conditions have been built into our neural network discretization. Thus, no penalty-type terms appear in the loss function. The lack of boundary terms in the loss may explain the simple loss landscapes we observed. Future work should focus on a comparison of techniques for boundary condition enforcement from the perspective of loss landscape visualization. Additionally, a loss landscape perspective on the spectral bias would provide important insight into one of the chief challenges of physics-informed learning.



\end{document}